\documentclass[10pt,twocolumn,letterpaper]{article}

\usepackage[pagenumbers]{cvpr} 

\usepackage{graphicx}
\usepackage{amsmath}
\usepackage{amssymb}
\usepackage{booktabs}
\usepackage{color}
\usepackage{nicefrac}

\usepackage{etoolbox}
\makeatletter
\pretocmd{\chapter}{\addtocontents{toc}{\protect\addvspace{15\p@}}}{}{}
\pretocmd{\section}{\addtocontents{toc}{\protect\addvspace{5\p@}}}{}{}
\pretocmd{\subsection}{\addtocontents{toc}{\protect\addvspace{3\p@}}}{}{}
\makeatother
\usepackage{titletoc}

\usepackage{xpatch}
\usepackage{array}
\usepackage{multirow}
\usepackage{graphicx}
\usepackage{xcolor,colortbl}
\usepackage{pifont}
\usepackage{enumitem}
\usepackage{xcolor}
\PassOptionsToPackage{dvipsnames}{xcolor}
\newcommand{\cmark}{\ding{51}}

\definecolor{Gray}{gray}{0.9}
\newcommand{\highlight}{\cellcolor{Gray}}
\definecolor{indian red}{RGB}{205,92,92}
\newcommand{\algname}{{{UniAD}}\xspace}
\usepackage{comment}
\usepackage[accsupp]{axessibility}

\definecolor{mycitecolor}{rgb}{0, 0.4, 0.7}
\usepackage[pagebackref,breaklinks,colorlinks,bookmarks,citecolor=mycitecolor]{hyperref}

\usepackage[capitalize]{cleveref}
\crefname{section}{Sec.}{Secs.}
\Crefname{section}{Section}{Sections}
\Crefname{table}{Table}{Tables}
\crefname{table}{Tab.}{Tabs.}

\def\cvprPaperID{1655} 
\def\confName{CVPR}
\def\confYear{2023}

\begin{document}

\title{
Planning-oriented
Autonomous Driving
}

\author{
Yihan Hu$^{1,2\ast}$,
Jiazhi Yang$^{1\ast}$,
Li Chen$^{1\ast \dagger}$,
Keyu Li$^{1\ast}$,
Chonghao Sima$^1$,
Xizhou Zhu$^{3,1}$ \\
Siqi Chai$^2$,
Senyao Du$^2$,
Tianwei Lin$^2$,
Wenhai Wang$^1$,
Lewei Lu$^3$,
Xiaosong Jia$^{1}$ \\
Qiang Liu$^2$,
Jifeng Dai$^1$,
Yu Qiao$^1$,
Hongyang Li$^{1\dagger}$ \\
[2mm]
$^1$~OpenDriveLab and OpenGVLab, Shanghai AI Laboratory \\ 
$^2$~Wuhan University \quad
$^3$~SenseTime Research
\\ 
\normalsize{
$^\ast$Equal contribution \quad $^\dagger$Project lead}
\\
\normalsize{
\url{https://github.com/OpenDriveLab/UniAD}
}
}

\maketitle


\begin{abstract}

Modern autonomous driving system is characterized as modular tasks in sequential order, i.e., perception, prediction, and planning. 
In order to perform a wide diversity of tasks and achieve advanced-level intelligence, contemporary approaches either deploy standalone models for individual tasks, or design a multi-task paradigm with separate heads. 
However, they might suffer from accumulative errors or deficient task coordination. Instead, we argue that a favorable framework should be devised and optimized in pursuit of the ultimate goal, i.e., planning of the self-driving car. 
Oriented at this, we revisit the key components within perception and prediction, and prioritize the tasks such that all these tasks contribute to planning. 
We introduce Unified Autonomous Driving (\algname), a comprehensive framework up-to-date that incorporates full-stack driving tasks in one network.
It is exquisitely devised to leverage advantages of each module, and provide complementary feature abstractions for agent interaction from a global perspective. Tasks are communicated with unified query interfaces to facilitate each other toward planning.
We instantiate \algname on the challenging nuScenes benchmark. With extensive ablations, the effectiveness of using such a philosophy is proven by substantially outperforming previous state-of-the-arts in all aspects. Code and models are public.
\end{abstract}

\section{Introduction}
\label{sec:intro}
\begin{figure}[t!]
  \centering
  \includegraphics[width=\linewidth]{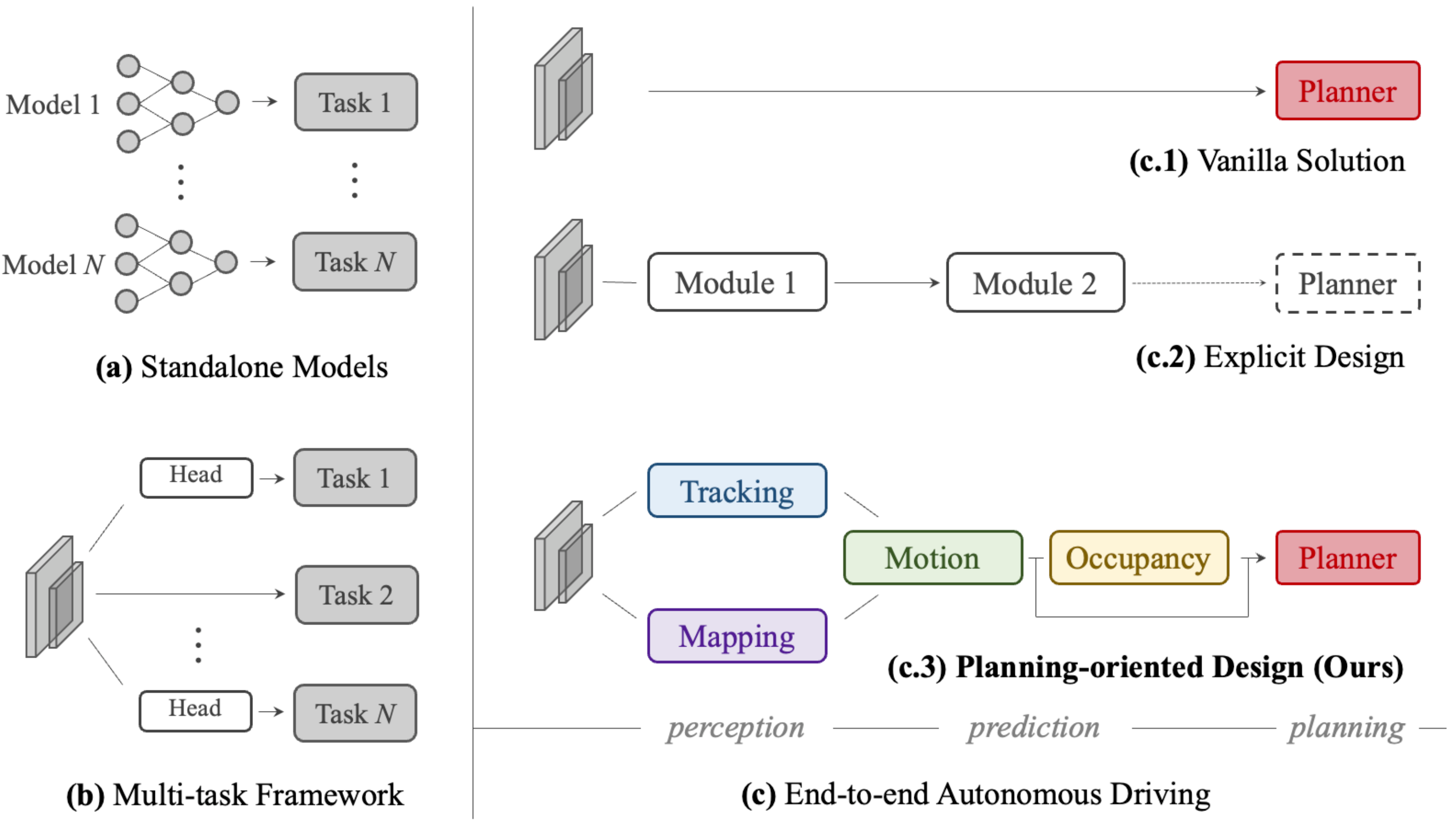}
  \vspace{-15pt}
  \caption{
  \textbf{Comparison on the various designs} of autonomous driving framework. \textbf{(a)} Most industrial solutions deploy separate models for different tasks. \textbf{(b)} The multi-task learning scheme shares a backbone with divided task heads.
  \textbf{(c)} The end-to-end paradigm unites modules in
  perception and prediction. Previous attempts either adopt a direct optimization on planning in (c.1) or devise the system with partial components in (c.2). Instead, we argue in (c.3) that a desirable system should be {planning-oriented} as well as properly organize preceding tasks to facilitate planning.
  }
  \label{fig:motivation}
  \vspace{-5pt}
\end{figure}

\vspace{-4pt}
With the successful development of deep learning, autonomous driving algorithms are assembled with a series of tasks\footnote{In the following context, we interchangeably use task, module, component, unit and node to indicate a certain task (\eg, detection).},
including detection, tracking, mapping in perception; and motion and occupancy forecast in prediction.
As depicted in~\cref{fig:motivation}(a), most industry solutions deploy standalone models for each task independently~\cite{mobileye2022ces,nvidia2022drive}, as long as the resource bandwidth of the onboard chip allows.
Although such a design simplifies the R\&D difficulty across teams, it bares the risk of information loss across modules, error accumulation and feature misalignment due to the isolation of optimization targets~\cite{luo2018faf,liang2020pnpnet,sadat2020p3}.

A more elegant design is to incorporate a wide span of tasks into a multi-task learning (MTL) paradigm, by plugging several task-specific heads into a shared feature extractor as shown in~\cref{fig:motivation}(b).
This is a popular practice in many domains, including general vision~\cite{wang2022ofa,zhu2022uniperceivermoe,reed2022gato}, autonomous driving\footnote{In this paper, we refer to MTL in autonomous driving as tasks \textit{beyond} perception. There is plenty of work on MTL \textit{within} perception, \eg, detection, depth, flow, \etc. This kind of literature is out of scope.}~\cite{zhang2022beverse,liang2022effective,zeng2019nmp,chen2022lav}, such as Transfuser~\cite{Chitta2022transfuserpami}, BEVerse~\cite{zhang2022beverse}, and industrialized products, \eg, Mobileye~\cite{mobileye2022ces}, Tesla~\cite{tesla2022aiday}, Nvidia~\cite{nvidia2022drive}, \etc.
In MTL, the co-training strategy across tasks could leverage feature abstraction; it could effortlessly extend to additional tasks, and save computation cost for onboard chips. 
However, such a scheme may cause undesirable ``negative transfer''~\cite{crawshaw2020mtlsurvey, liu2022bevfusion}.

By contrast, the emergence of end-to-end autonomous driving~\cite{chitta2021neat,wu2022tcp,chen2022lav,casas2021mp3,hu2022stp3} unites all nodes from perception, prediction and planning as a \textit{whole}.
The choice and priority of preceding tasks should be determined in favor of planning.
The system should be planning-oriented, exquisitely designed with certain components involved, such that there are few accumulative error as in the standalone option or negative transfer as in the MTL scheme. \Cref{tab:motivation} describes the task taxonomy of different framework designs.

Following the end-to-end paradigm, one ``tabula-rasa'' practice is to directly predict the planned trajectory, without any explicit supervision of perception and prediction as shown in~\cref{fig:motivation}(c.1).
Pioneering works~\cite{chen2021wor,codevilla2018cil, codevilla2019cilrs, chen2020lbc, zhang2021roach, prakash2021transfuser, wu2022tcp, wu2023ppgeo} verified this vanilla design in the closed-loop simulation~\cite{Dosovitskiy17carla}. While such a direction deserves further exploration, it is inadequate in safety guarantee and interpretability, especially for highly dynamic urban scenarios.
In this paper, we lean toward another perspective and ask the following question:
\emph{Toward a reliable and planning-oriented autonomous driving system, how to design the pipeline in favor of planning? which preceding tasks are requisite?}

\begin{table}[t!]
    \centering
    \scalebox{0.7}{
	\begin{tabular}{l|l|ccc|cc|c}
		\toprule
		\multirow{2}{*}{Design} & \multirow{2}{*}{Approach} & \multicolumn{3}{c|}{Perception} & \multicolumn{2}{c|}{Prediction} & \multirow{2}{*}{Plan} \\
		& & Det. & Track & Map & Motion & Occ. & \\
		\midrule
		\multirow{3}{*}{(b)} & NMP~\cite{zeng2019nmp} & \cmark & & & \cmark & & \cmark \\
		 & NEAT~\cite{chitta2021neat} &  &  & \cmark &  &  & \cmark \\
		 & BEVerse~\cite{zhang2022beverse} & \cmark & & \cmark & & \cmark & \\
		\midrule
		(c.1) & \cite{chen2021wor,chen2020lbc,prakash2021transfuser,wu2022tcp} &  &  &  & & & \cmark\\
		\midrule
		\multirow{6}{*}{(c.2)} & PnPNet$^\text{\textdagger}$~\cite{liang2020pnpnet} & \cmark & \cmark &  & \cmark & & \\
		 & ViP3D$^\text{\textdagger}$~\cite{gu2022vip3d} & \cmark & \cmark &  & \cmark & & \\
		 & P3~\cite{sadat2020p3} &  &  &  &  & \cmark & \cmark \\
		 & MP3~\cite{casas2021mp3} &  &  & \cmark &  & \cmark & \cmark \\
		 & ST-P3~\cite{hu2022stp3} &  &  & \cmark & & \cmark & \cmark \\
		 & LAV~\cite{chen2022lav} & \cmark &  & \cmark & \cmark &  & \cmark \\
		\midrule
		(c.3) & \textbf{\algname} (ours) & \cmark & \cmark & \cmark & \cmark & \cmark & \cmark \\
		\bottomrule
	\end{tabular}
	}
 \vspace{-5pt}
	\caption{\textbf{Tasks comparison and taxonomy.} ``Design'' column is classified as in~\cref{fig:motivation}. ``Det.'' denotes 3D object detection, ``Map'' stands for online mapping, and ``Occ.'' is occupancy map prediction. \textdagger: these works are not proposed directly for planning, yet they still share the spirit of joint perception and prediction. \algname conducts five essential driving tasks to facilitate planning.
 }
	\label{tab:motivation}
\end{table} 

An intuitive resolution would be to perceive surrounding objects, predict future behaviors and plan a safe maneuver explicitly, as illustrated in~\cref{fig:motivation}(c.2). 
Contemporary approaches~\cite{liang2020pnpnet, gu2022vip3d, sadat2020p3, casas2021mp3, hu2022stp3} provide good insights and achieve impressive performance. However, we argue that the devil lies in the details; previous works more or less fail to consider certain components (see block (c.2) in~\Cref{tab:motivation}), being reminiscent of the planning-oriented spirit. 
We elaborate on the detailed definition and terminology, the necessity of these modules in the Supplementary.

To this end, we introduce \textbf{\algname}, a Unified Autonomous Driving algorithm framework to leverage five essential tasks toward a safe and robust system as depicted in~\cref{fig:motivation}(c.3) and~\Cref{tab:motivation}(c.3).
\algname is designed in a planning-oriented spirit. We argue that this is \textit{not} a simple stack of tasks with mere engineering effort.
A key component is the query-based design to connect all nodes. Compared to the classic bounding box representation, queries benefit from a larger receptive field to soften the compounding error from upstream predictions. Moreover, queries are flexible to model and encode a variety of interactions, \eg, relations among multiple agents. 
To the best of our knowledge, \algname is the first work to comprehensively investigate the joint cooperation of such a variety of tasks including perception, prediction and planning in the field of autonomous driving.

The contributions are summarized as follows. 
\textbf{(a)} we embrace a new outlook of autonomous driving framework following a planning-oriented philosophy, and demonstrate the necessity of effective task coordination, rather than standalone design or simple multi-task learning. 
\textbf{(b)} we present \algname, a comprehensive end-to-end system that leverages a wide span of tasks. The key component to hit the ground running is the query design as interfaces connecting all nodes. 
As such, \algname enjoys flexible intermediate representations and exchanging multi-task knowledge toward planning. 
\textbf{(c)} we instantiate \algname on the challenging benchmark for realistic scenarios. Through extensive ablations, we verify the superiority of our method over previous state-of-the-arts in all aspects.

\emph{We hope this work could shed some light on the target-driven design for the autonomous driving system, providing a starting point for coordinating various driving tasks.}

\begin{figure*}[th!]
  \centering
  \includegraphics[width=\linewidth]{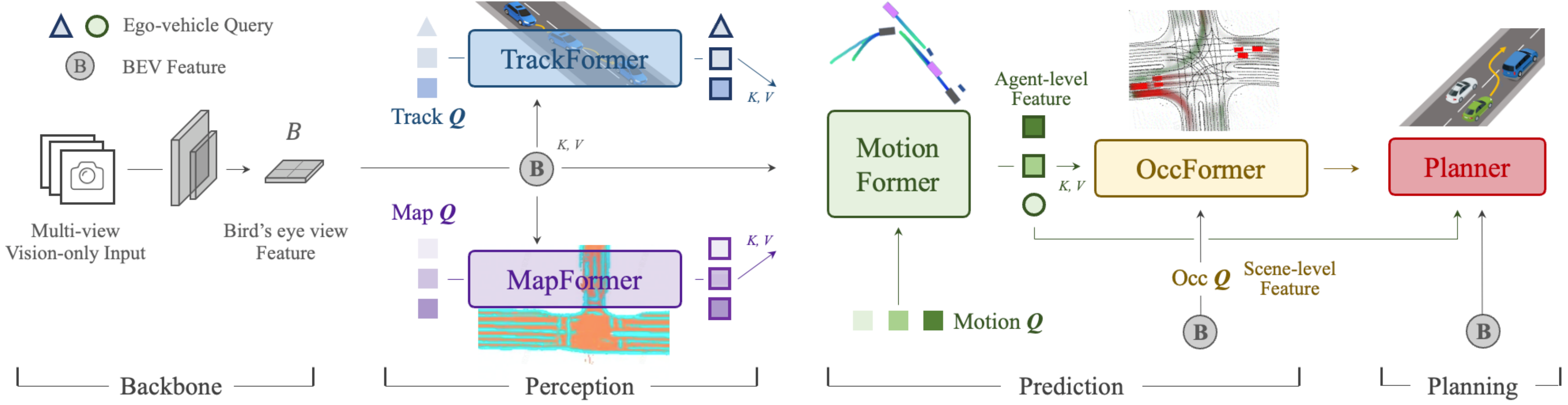}
  \vspace{-15pt}
  \caption{
  \textbf{Pipeline} of Unified Autonomous Driving (\algname). It is exquisitely devised following planning-oriented philosophy. Instead of a simple stack of tasks, we investigate the effect of each module in perception and prediction, leveraging the benefits of joint optimization from preceding nodes to final planning in the driving scene.
  All perception and prediction modules are designed in a transformer decoder structure, with task queries as interfaces connecting each node.
  A simple attention-based planner is in the end to predict future waypoints of the ego-vehicle considering the knowledge extracted from preceding nodes. The map over occupancy is for visual purpose only.
  }
  \label{fig:pipeline}
\end{figure*}

\section{Methodology}
\label{sec:method}
\vspace{-4pt}

\paragraph{Overview.} As illustrated in~\cref{fig:pipeline}, \algname comprises four transformer decoder-based perception and prediction modules and one planner in the end. Queries $Q$ play the role of connecting the pipeline to model different interactions of entities in the driving scenario.
Specifically, a sequence of multi-camera images is fed into the feature extractor, and the resulting perspective-view features are transformed into a unified bird's-eye-view (BEV) feature $B$ by an off-the-shelf BEV encoder in BEVFormer~\cite{li2022bevformer}. Note that \algname is not confined to a specific BEV encoder, and one can utilize other alternatives to extract richer BEV representations with long-term temporal fusion~\cite{han2023exploring, park2022solofusion} or multi-modality fusion~\cite{liu2022bevfusion, liang2022bevfusion}.
In \textbf{TrackFormer}, the learnable embeddings that we refer to as track queries inquire about the agents' information from $B$ to detect and track agents.
\textbf{MapFormer} takes map queries as semantic abstractions of road elements (\eg, lanes and dividers) and performs panoptic segmentation of the map.
With the above queries representing agents and maps, \textbf{MotionFormer} captures interactions among agents and maps and forecasts per-agent future trajectories.
Since the action of each agent can significantly impact others in the scene, this module makes joint predictions for all agents considered.
Meanwhile, we devise an ego-vehicle query to explicitly model the ego-vehicle and enable it to interact with other agents in such a scene-centric paradigm. 
\textbf{OccFormer} employs the BEV feature $B$ as queries, equipped with agent-wise knowledge as keys and values, and predicts multi-step future occupancy with agent identity preserved.
Finally, \textbf{Planner} utilizes the expressive ego-vehicle query from MotionFormer to predict the planning result, and keep itself away from occupied regions predicted by OccFormer to avoid collisions.

\subsection{Perception: Tracking and Mapping}
\vspace{-4pt}
\paragraph{TrackFormer.} It jointly performs detection and multi-object tracking (MOT) without non-differentiable post-processing. Inspired by~\cite{zeng2021motr, zhang2022mutr3d}, we take a similar query design.
Besides the conventional detection queries utilized in object detection~\cite{carion2020detr, zhu2020deformabledetr}, additional track queries are introduced to track agents across frames.
Specifically, at each time step, initialized detection queries are responsible for detecting newborn agents that are perceived for the first time, while track queries keep modeling those agents detected in previous frames. Both detection queries and track queries capture the agent abstractions by attending to BEV feature $B$.
As the scene continuously evolves, track queries at the current frame interact with previously recorded ones in a self-attention module to aggregate temporal information, until the corresponding agents disappear completely (untracked in a certain time period).
Similar to~\cite{carion2020detr}, TrackFormer contains $N$ layers and the final output state $Q_A$ provides knowledge of $N_a$ valid agents for downstream prediction tasks.
Besides queries encoding other agents surrounding the ego-vehicle, we introduce one particular \emph{ego-vehicle query} in the query set to explicitly model the self-driving vehicle itself, which is further used in planning.

\paragraph{MapFormer.} We design it based on a 2D panoptic segmentation method Panoptic SegFormer~\cite{li2022panopticseg}. 
We sparsely represent road elements as map queries to help downstream motion forecasting, with location and structure knowledge encoded. For driving scenarios, we set lanes, dividers and crossings as things, and the drivable area as stuff~\cite{kirillov2019panoptic}.
MapFormer also has $N$ stacked layers whose output results of each layer are all supervised, while only the updated queries $Q_M$ in the last layer are forwarded to MotionFormer for agent-map interaction.

\subsection{Prediction: Motion Forecasting}
\label{sec:motion-method}
\vspace{-4pt}
Recent studies have proven the effectiveness of transformer structure on the motion task~\cite{liu2021multimodal, yuan2021agentformer, jia2021allocentric, jia2022hdgt, nayakanti2022wayformer, shi2022motiontransformer, ngiam2021scenetransformer}, inspired by which we propose MotionFormer in the end-to-end setting. 
With highly abstract queries for dynamic agents $Q_A$ and static map $Q_M$ from TrackFormer and MapFormer respectively, MotionFormer predicts all agents' multimodal future movements, \ie, top-k possible trajectories, in a scene-centric manner.
This paradigm produces multi-agent trajectories in the frame with a single forward pass, which greatly saves the computational cost of aligning the whole scene to each agent's coordinate~\cite{kim2022stopnet}.
Meanwhile, we pass the \emph{ego-vehicle query} from TrackFormer through MotionFormer to engage ego-vehicle to interact with other agents, considering the future dynamics.
Formally, the output motion is formulated as 
$\!\{\hat{\mathbf{x}}_{i,k}\!\in\!\mathbb{R}^{T\!\times\!2} | i\!=1,\dots,N_a; k\!=\!1,\dots,\mathcal{K}\}$
, where $i$ indexes the agent, $k$ indexes the modality of trajectories and $T$ is the length of prediction horizon.

\paragraph{MotionFormer.} It is composed of $N$ layers, and each layer captures three types of interactions: agent-agent, agent-map and agent-goal point.
For each motion query $Q_{i,k}$ (defined later, and we omit subscripts $i$,\,$k$ in the following context for simplicity), its interactions between other agents $Q_{A}$ or map elements $Q_{M}$ could be formulated as:
\begin{equation}
\label{eq:agent-agent}
    Q_{a/m} = \texttt{MHCA}(\texttt{MHSA}(Q), Q_A/Q_M),
\end{equation}
where $\texttt{MHCA}$, $\texttt{MHSA}$ denote multi-head cross-attention and multi-head self-attention~\cite{vaswani2017transformer} respectively.
As it is also important to focus on the intended position, \ie, goal point, to refine the predicted trajectory, we devise an agent-goal point attention via deformable attention~\cite{zhu2020deformabledetr} as follows:
\begin{equation}
\label{eq:agent-bev}
    Q_{g} = \texttt{DeformAttn}(Q, \hat{\mathbf{x}}_{T}^{l-1}, B),
\end{equation}
where $\hat{\mathbf{x}}_T^{l-1}$ is the endpoint of the predicted trajectory of previous layer. $\texttt{DeformAttn}(q,\!r,\!x)$, a deformable attention module, takes in the query $q$, reference point $r$ and spatial feature $x$. 
It performs sparse attention on the spatial feature around the reference point. Through this, the predicted trajectory is further refined as aware of the endpoint surroundings.
All three interactions are modeled in parallel, where the generated $Q_a$, $Q_m$ and $Q_g$ are concatenated and passed to a multi-layer perceptron (MLP), resulting query context $Q_{\text{ctx}}$. Then, $Q_{\text{ctx}}$ is sent to the successive layer for refinement or decoded as prediction results at the last layer.
\paragraph{Motion queries.} The input queries for each layer of MotionFormer, termed motion queries, comprise two components: the query context $Q_{\text{ctx}}$ produced by the preceding layer as described before, and the query position $Q_{\text{pos}}$.
Specifically, $Q_{\text{pos}}$ integrates the positional knowledge in four-folds as in~\cref{eq:qpos}: (1) the position of scene-level anchors $I^s$; (2) the position of agent-level anchors $I^a$; (3) current location of the agent $i$ and (4) the predicted goal point.
\begin{equation}
\label{eq:qpos}
\begin{aligned}
    Q_{\text{pos}} =\ &\text{MLP}(\text{PE}(I^s))  + \text{MLP}(\text{PE}(I^a))  \\
    +\ &\text{MLP}(\text{PE}(\hat{\mathbf{x}}_0)) + \text{MLP}(\text{PE}(\hat{\mathbf{x}}_T^{l-1})).
\end{aligned}
\end{equation}
Here the sinusoidal position encoding $\text{PE}(\cdot)$ followed by an MLP is utilized to encode the positional points and $\hat{\mathbf{x}}_T^0$ is set as $I^s$ at the first layer (subscripts $i,k$ are also omitted).
The scene-level anchor represents prior movement statistics in a global view, while the agent-level anchor captures the possible intention in the local coordinate. They are both clustered by k-means algorithm on the endpoints of ground-truth trajectories, to narrow down the uncertainty of prediction.
Contrary to the prior knowledge, the start point provides customized positional embedding for each agent, and the predicted endpoint serves as a dynamic anchor optimized layer-by-layer in a coarse-to-fine fashion.

\paragraph{Non-linear Optimization.}
Different from conventional motion forecasting works which have direct access to ground truth perceptual results, \ie, agents' location and corresponding tracks, we consider the prediction uncertainty from the prior module in our end-to-end paradigm.
Brutally regressing the ground-truth waypoints from an imperfect detection position or heading angle may lead to unrealistic trajectory predictions with large curvature and acceleration.
To tackle this, we adopt a non-linear smoother~\cite{nuplan} to adjust the target trajectories and make them physically feasible given an imprecise starting point predicted by the upstream module. The process is:
\begin{equation}
\label{eq:non-linear-argmin}
    \tilde{\mathbf{x}}^* = \arg \min _{\mathbf{x}} c(\mathbf{x}, \tilde{\mathbf{x}} ),
\end{equation}
where $\tilde{\mathbf{x}}$ and $\tilde{\mathbf{x}}^*$ denote the ground-truth and smoothed trajectory, $\mathbf{x}$ is generated by multiple-shooting~\cite{bock1984multiple}, and the cost function is as follows:
\begin{equation}
\label{eq:non-linear-cost}
    c(\mathbf{x}, \tilde{\mathbf{x}} ) = \lambda_{\text{xy}} \lVert\mathbf{x}, \tilde{\mathbf{x}}\rVert_2 +  \lambda_{\text{goal}} \lVert\mathbf{x}_{T}, \tilde{\mathbf{x}}_{T}\rVert_2 
    +  \sum_{\phi\in \Phi} \phi(\mathbf{x}),
\end{equation}
where $\lambda_{\text{xy}}$ and $\lambda_{\text{goal}}$ are hyperparameters, the kinematic function set $\Phi$ has five terms including jerk, curvature, curvature rate, acceleration and lateral acceleration. The cost function regularizes the target trajectory to obey kinematic constraints. This target trajectory optimization is only conducted in training and does not affect inference.

\subsection{Prediction: Occupancy Prediction}
\label{sec:occ-method}
\vspace{-4pt}

Occupancy grid map is a discretized BEV representation where each cell holds a belief indicating whether it is occupied, and the occupancy prediction task is to discover how the grid map changes in the future.
Previous approaches utilize RNN structure for temporally expanding future predictions from observed BEV features~\cite{hu2021fiery, hu2022stp3, zhang2022beverse}.
However, they rely on highly hand-crafted clustering post-processing to generate per-agent occupancy maps, as they are mostly agent-agnostic by compressing BEV features as a whole into RNN hidden states.
Due to the deficient usage of agent-wise knowledge, it is challenging for them to predict the behaviors of all agents globally, which is essential to understand how the scene evolves.
To address this, we present OccFormer to incorporate both scene-level and agent-level semantics in two aspects: (1) a dense scene feature acquires agent-level features via an exquisitely designed attention module when unrolling to future horizons; (2) we produce instance-wise occupancy easily by a matrix multiplication between agent-level features and dense scene features without heavy post-processing.

OccFormer is composed of $T_o$ sequential blocks where $T_o$ indicates the prediction horizon. Note that $T_o$ is typically smaller than $T$ in the motion task, due to the high computation cost of densely represented occupancy.
Each block takes as input the rich agent features $G^t$ and the state (dense feature) $F^{t-1}$ from the previous layer, and generates $F^t$ for timestep $t$ considering both instance- and scene-level information.
To get agent feature $G^t$ with dynamics and spatial priors, 
we max-pool motion queries from MotionFormer in the modality dimension denoted as $Q_X\!\in\!\mathbb{R}^{N_a\!\times\!D}$, with $D$ as the feature dimension. Then we fuse it with the upstream track query $Q_{A}$ and current position embedding $P_{A}$ via a temporal-specific MLP:
\begin{equation}
\label{eq:query fuse}
    G^t = \text{MLP}_{t}([Q_{A}, P_{A}, Q_{X}]),\ t={1,\dots,T_o},
\end{equation}
where $[\cdot]$ indicates concatenation. For the scene-level knowledge, the BEV feature $B$ is downscaled to \nicefrac{1}{4} resolution for training efficiency to serve as the first block input $F^0$.
To further conserve training memory, each block follows a downsample-upsample manner with an attention module in between to conduct pixel-agent interaction at \nicefrac{1}{8} downscaled feature, denoted as $F_{\text{ds}}^t$.

\vspace{-7pt}
\paragraph{Pixel-agent interaction} is designed to unify the scene- and agent-level understanding when predicting future occupancy.
We take the dense feature $F_\text{ds}^t$ as queries, instance-level features as keys and values to update the dense feature over time. Detailedly, $F^t_{\text{ds}}$ is passed through a self-attention layer to model responses between distant grids, then a cross-attention layer models interactions between agent features ${G^{t}}$ and per-grid features.
Moreover, to align the pixel-agent correspondence, we constrain the cross-attention by an attention mask, which restricts each pixel to only look at the agent occupying it at timestep $t$, inspired by~\cite{cheng2021mask2former}.
The update process of the dense feature is formulated as:
\begin{equation}
\label{eq:mask attention}
    D_{\text{ds}}^t = \texttt{MHCA}( \texttt{MHSA}(F_{\text{ds}}^t), G^t, \text{attn\_mask} = O_m^t).
\end{equation}
The attention mask $O_m^t$ is semantically similar to occupancy, and is generated by multiplying an additional agent-level feature and the dense feature $F_{\text{ds}}^t$, where we name the agent-level feature here as mask feature $M^t=\text{MLP}({G^{t}})$. 
After the interaction process in~\cref{eq:mask attention}, $D_{\text{ds}}^t$ is upsampled to \nicefrac{1}{4} size of $B$. We further add $D_{\text{ds}}^t$ with block input $F^{t-1}$ as a residual connection, and the resulting feature $F^t$ is passed to the next block. 

\vspace{-7pt}
\paragraph{Instance-level occupancy.} It represents the occupancy with each agent's identity preserved. It could be simply drawn via matrix multiplication, as in recent query-based segmentation works~\cite{cheng2021maskformer, li2022maskdino}.
Formally, in order to get an occupancy prediction of original size $H\!\times\!W$ of BEV feature $B$, the scene-level features $F^t$ are upsampled to $F_{\text{dec}}^t\!\in\!\mathbb{R}^{C\!\times\!H\!\times\!W}$ by a convolutional decoder, where $C$ is the channel dimension.
For the agent-level feature, we further update the coarse mask feature $M^t$ to the occupancy feature $U^t\!\in\!\mathbb{R}^{N_a\!\times\!C}$ by another MLP. We empirically find that generating $U^t$ from mask feature $M^t$ instead of original agent feature ${G^{t}}$ leads to superior performance.
The final instance-level occupancy of timestep $t$ is:
\begin{equation}
    \hat{O}_A^t = U^t \cdot F_{\text{dec}}^t.
\end{equation}

\subsection{Planning}
\label{sec:method-plan}
\vspace{-4pt}
Planning without high-definition (HD) maps or predefined routes usually requires a high-level command to indicate the direction to go~\cite{casas2021mp3, hu2022stp3}. 
Following this, we convert the raw navigation signals (\ie, turn left, turn right and keep forward) into three learnable embeddings, named command embeddings.
As the ego-vehicle query from MotionFormer already expresses its multimodal intentions, we equip it with command embeddings to form a ``plan query". We attend plan query to BEV features $B$ to make it aware of surroundings, and then decode it to future waypoints $\hat{\tau}$.  

To further avoid collisions, we optimize $\hat{\tau}$ based on Newton's method in inference only by the following:
\begin{equation}
\label{eq:plan-argmin}
    \tau^* = \arg \min _{\tau} f(\tau, \hat{\tau}, \hat{O} ),
\end{equation}
where $\hat{\tau}$ is the original planning prediction, $\tau^*$ denotes the optimized planning, which is selected from multiple-shooting \cite{bock1984multiple} trajectories $\tau$ as to minimize cost function $f(\cdot)$. $\hat{O}$ is a classical binary occupancy map merged from the instance-wise occupancy prediction from OccFormer. The cost function $f(\cdot)$ is calculated by:
\begin{equation}
\label{eq:col-cost}
    f(\tau, \hat{\tau}, \hat{O} ) = \lambda_{\text{coord}}  \lVert\tau, \hat{\tau}\rVert_2 +  
    \lambda_{\text{obs}}  \sum_{t} \mathcal{D}(\tau_{t}, \hat{O}^{t}), 
\end{equation}
\vspace{-10pt}
\begin{equation}
\label{eq:col-dist}
    \mathcal{D}(\tau_{t}, \hat{O}^{t}) = \sum_{(x, y) \in \mathcal{S}} 
    \frac{1}{\sigma \sqrt{2\pi}}
    \text{exp} (-\frac{\lVert \tau_{t} - (x, y) \rVert_2^2}{2\sigma^2}). 
\end{equation}
Here $\lambda_{\text{coord}}$, $\lambda_{\text{obs}}$, and $\sigma$ are hyperparameters, and $t$ indexes a timestep of future horizons. The $l_2$ cost pulls the trajectory toward the original predicted one, while the collision term $\mathcal{D}$ pushes it away from occupied grids, considering surrounding positions confined to $ \mathcal{S}=\{(x, y) | \lVert(x, y) - \tau_t\rVert_{2} < d, \hat{O}_{x, y}^t = 1\}$.

\begin{table*}[t!]
    \definecolor{Gray}{gray}{0.9}
	\begin{center}
		\resizebox{\textwidth}{!}{
			\begin{tabular}{l|ccccc|ccc|cc|ccc|cccc|cc}
				\toprule
				\multirow{2}{*}{ID} &
				\multicolumn{5}{c|}{Modules} & 
				\multicolumn{3}{c|}{Tracking} & 
				\multicolumn{2}{c|}{Mapping} & 
				\multicolumn{3}{c|}{Motion Forecasting} & 
				\multicolumn{4}{c|}{Occupancy Prediction} & \multicolumn{2}{c}{Planning}  \\
				& Track & Map & Motion & Occ. & Plan& AMOTA$\uparrow$ & AMOTP$\downarrow$ &IDS$\downarrow$ & IoU-lane$\uparrow$ & IoU-road$\uparrow$ & minADE$\downarrow$ & minFDE$\downarrow$ & MR$\downarrow$ & IoU-n.$\uparrow$& IoU-f.$\uparrow$ & VPQ-n.$\uparrow$ & VPQ-f.$\uparrow$& avg.L2$\downarrow$  & avg.Col.$\downarrow$   \\
				\midrule
				0$^\ast$ & \cmark & \cmark & \cmark & \cmark & \cmark & 0.356 & 1.328 &893  & 0.302 & 0.675 & 0.858 & 1.270 & 0.186 & 55.9 & 34.6 & 47.8 &26.4  & 1.154 & 0.941 \\
				\midrule
				1 & \cmark & & & & & \highlight{0.348} & \highlight{1.333} & \highlight{791} & \highlight{-} & \highlight{-} & - & - & - & - & - & - & - & - & - \\
				2 & & \cmark & & & & \highlight{-} & \highlight{-} & \highlight{-} & \highlight{\textbf{0.305}} & \highlight{\underline{0.674}} & - & - & - & - & - & - & - & - & - \\
				3 & \cmark & \cmark & & & & \highlight{0.355} & \highlight{1.336} & \highlight{\underline{785}} & \highlight{0.301} & \highlight{0.671} & - & - & - & - & - & - & - & - & - \\
				\midrule
				4 &  &  & \cmark &  &  & - & - & - & - & - & \highlight{0.815} & \highlight{1.224} & \highlight{0.182} & - & - & - & - & - & - \\
				5 & \cmark &  & \cmark &  &  & \underline{0.360} & 1.350 & 919 & - & - & \highlight{0.751} & \highlight{1.109} & \highlight{0.162} & - & - & - & - & - & - \\
				6 & \cmark & \cmark & \cmark &  &  & 0.354 & 1.339 & 820 & 0.303 & 0.672 & \highlight{0.736(-9.7\%)} & \highlight{1.066(-12.9\%)} & \highlight{0.158} & - & - & - & - & - & - \\
				\midrule
				7 &  &  &  & \cmark &  & - & - & - & - & - & \highlight{-} & \highlight{-} & \highlight{-} & \highlight{60.5}& \highlight{37.0}& \highlight{52.4}	&\highlight{29.8}  & - & - \\
				8 & \cmark &  &  & \cmark &  & \underline{0.360} & \textbf{1.322} & 809 & - & - & \highlight{-} & \highlight{-} & \highlight{-} & \highlight{\underline{62.1}} & \highlight{38.4} & \highlight{52.2} & \highlight{32.1}  & - & - \\
				9 & \cmark & \cmark & \cmark & \cmark & & 0.359 & 1.359 & 1057 & \underline{0.304} & \textbf{0.675} & \highlight{\textbf{0.710}(-3.5\%)} & \highlight{\textbf{1.005}(-5.8\%)} & \highlight{\textbf{0.146}} &\highlight{\textbf{62.3}} & \highlight{\underline{39.4}} & \highlight{\textbf{53.1}} & \highlight{\underline{32.2}} & - & - \\
				\midrule
				10 &  &  &  &  & \cmark &  & - & - & - & - & - & - & - & - & - & - & - & \highlight{1.131} & \highlight{0.773} \\
				11 & \cmark & \cmark & \cmark &  & \cmark &  \textbf{0.366} & 1.337 & 889 & 0.303 & 0.672 & 0.741 & 1.077 & 0.157&  - & - & - & - & \highlight{\underline{1.014}} & \highlight{\underline{0.717}} \\
				12 & \cmark & \cmark & \cmark & \cmark & \cmark & 0.358	 & \underline{1.334} & \textbf{641} & 0.302 & 0.672 & \underline{0.728} & \underline{1.054} & \underline{0.154} & \textbf{62.3} & \textbf{39.5} & \underline{52.8} & \textbf{32.3} & \highlight{\textbf{1.004}} & \highlight{\textbf{0.430}} \\
				\bottomrule
			\end{tabular}
		}
	\end{center}
 \vspace{-15pt}
	\caption{\textbf{Detailed ablations on the effectiveness of each task.} We can conclude that two perception sub-tasks greatly help motion forecasting, and prediction performance also benefits from unifying the two prediction modules. With all prior representations, our goal-planning boosts significantly to ensure safety.
 \algname outperforms naive MTL solution by a large margin for prediction and planning tasks, and it also owns the superiority that \textit{no} substantial perceptual performance drop occurs. Only main metrics are shown for brevity.  
 ``avg.L2'' and ``avg.Col'' are the average values across the planning horizon. $\ast$: ID-0 is the MTL scheme with separate heads for each task.
 }
	\label{tab:abl-module}
\end{table*}

\subsection{Learning}
\vspace{-4pt}

\algname is trained in two stages. We first jointly train perception parts, \ie, the tracking and mapping modules, for a few epochs (6 in our experiments), and then train the model end-to-end for 20 epochs with all perception, prediction and planning modules. The two-stage training is found more stable empirically. We refer the audience to the Supplementary for details of each loss.

\vspace{-4pt}
\paragraph{Shared matching.} Since \algname involves instance-wise modeling, pairing predictions to the ground truth set is required in perception and prediction tasks.
Similar to DETR~\cite{carion2020detr,li2022panopticseg}, the bipartite matching algorithm is adopted in the tracking and online mapping stage. As for tracking, candidates from detection queries are paired with newborn ground truth objects, and predictions from track queries inherit the assignment from previous frames.
The matching results in the tracking module are reused in motion and occupancy nodes to consistently model agents from historical tracks to future motions in the end-to-end framework.

\section{Experiments}
\label{sec:experiments}
\vspace{-4pt}

We conduct experiments on the challenging nuScenes dataset~\cite{caesar2020nuscenes}. In this section, we validate the effectiveness of our design in three aspects: joint results revealing the advantage of task coordination and its effect on planning, modular results of each task compared with previous methods, and ablations on the design space for specific modules. Due to space limit, the full suite of protocols, some ablations and visualizations are provided in the Supplementary.

\subsection{Joint Results}
\label{sec:exp-joint}
\vspace{-4pt}
We conduct extensive ablations as shown in~\Cref{tab:abl-module} to prove the effectiveness and necessity of preceding tasks in the end-to-end pipeline. Each row of this table shows the model performance when incorporating task modules listed in the second \textit{Modules} column. 
The first row (ID-0) serves as a vanilla multi-task baseline with separate task heads for comparison.
The best result of each metric is marked in bold, and the runner-up result is underlined in each column.

\vspace{-4pt}
\paragraph{Roadmap toward safe planning.}
As prediction is closer to planning compared to perception, 
we first investigate the two types of prediction tasks in our framework, \ie, motion forecasting and occupancy prediction. In Exp.10-12, only when the two tasks are introduced simultaneously (Exp.12), both metrics of the planning L2 and collision rate achieve the best results, compared to naive end-to-end planning without any intermediate tasks (Exp.10, \cref{fig:motivation}(c.1)).
Thus we conclude that both these two prediction tasks are required for a safe planning objective.
Taking a step back, in Exp.7-9, we show the cooperative effect of two types of prediction. The performance of both tasks get improved when they are closely integrated (Exp.9, -3.5\%\,minADE, -5.8\%\,minFDE, -1.3 MR(\%), +2.4\,IoU-f.(\%), +2.4\,VPQ-f.(\%)), which demonstrates the necessity to include both agent and scene representations.
Meanwhile, in order to realize a superior motion forecasting performance, we explore how perception modules could contribute in Exp.4-6. Notably, incorporating both tracking and mapping nodes brings remarkable improvement to forecasting results (-9.7\%\,minADE, -12.9\%\,minFDE, -2.3 MR(\%)).
We also present Exp.1-3, which indicate training perception sub-tasks together leads to comparable results to a single task.
Additionally, compared with naive multi-task learning (Exp.0, \cref{fig:motivation}(b)), Exp.12 outperforms it by a significant margin in all essential metrics (-15.2\%\,minADE, -17.0\%\,minFDE, -3.2\,MR(\%)), +4.9\,IoU-f.(\%)., +5.9\,VPQ-f.(\%), -0.15$m$\,avg.L2, -0.51 avg.Col.(\%)), showing the superiority of our planning-oriented design.

\subsection{Modular Results}
\label{sec:exp-sota}
\vspace{-4pt}
Following the sequential order of perception-prediction-planning, we report the performance of each task module in comparison to prior state-of-the-arts on the nuScenes validation set. Note that \algname jointly performs all these tasks with a single trained network. The main metric for each task is marked with gray background in tables. 

\vspace{-6pt}
\paragraph{Perception results.} As for multi-object tracking in \Cref{tab:sota-track}, \algname yields a significant improvement of \textbf{+6.5} and \textbf{+14.2} AMOTA(\%) compared to MUTR3D~\cite{zhang2022mutr3d} and ViP3D~\cite{gu2022vip3d} respectively.
Moreover, \algname achieves the lowest ID switch score, 
 showing its temporal consistency for each tracklet. For online mapping in \Cref{tab:sota-map}, \algname performs well on segmenting lanes (\textbf{+7.4} IoU(\%) compared to BEVFormer), which is crucial for downstream agent-road interaction in the motion module.
As our tracking module follows an end-to-end paradigm, it is still inferior to tracking-by-detection methods with complex associations such as Immortal Tracker~\cite{wang2021immortal}, and our mapping results trail previous perception-oriented methods on specific classes. We argue that \algname is to benefit \textit{final} planning with perceived information rather than optimizing perception with full model capacity.

\vspace{-6pt}
\paragraph{Prediction results.} Motion forecasting results are shown in \Cref{tab:sota-motion}, where \algname remarkably outperforms previous vision-based end-to-end methods. It reduces prediction errors by \textbf{38.3\%} and \textbf{65.4\%} on minADE compared to PnPNet-vision~\cite{liang2020pnpnet} and ViP3D~\cite{gu2022vip3d} respectively.
In terms of occupancy prediction reported in \Cref{tab:sota-occ}, \algname gets notable advances in nearby areas, yielding \textbf{+4.0} and \textbf{+2.0} on IoU-near(\%) compared to FIERY~\cite{hu2021fiery} and BEVerse~\cite{zhang2022beverse} with heavy augmentations, respectively.

\vspace{-6pt}
\paragraph{Planning results.} Benefiting from rich spatial-temporal information in both the ego-vehicle query and occupancy, \algname reduces planning L2 error and collision rate by \textbf{51.2\%} and \textbf{56.3\%} compared to ST-P3~\cite{hu2022stp3}, in terms of the average value for the planning horizon.
Moreover, it notably outperforms several LiDAR-based counterparts, which is often deemed challenging for perception tasks.

\begin{table}[t!]
    \centering
    \scalebox{0.8}{
	\begin{tabular}{l|cccc}
		\toprule
		Method & \cellcolor{gray!30}AMOTA$\uparrow$ & AMOTP$\downarrow$ & Recall$\uparrow$ & IDS$\downarrow$ \\
		\midrule
		Immortal Tracker$^{\text{\textdagger}}$~\cite{wang2021immortal} & \cellcolor{gray!30}{0.378} & {1.119} & {0.478} & {936} \\
		\midrule
		ViP3D~\cite{gu2022vip3d} & \cellcolor{gray!30}0.217 & 1.625 & 0.363 & - \\
		QD3DT~\cite{hu2022qd3dt} & \cellcolor{gray!30}0.242 & 1.518 & 0.399 & - \\
		MUTR3D~\cite{zhang2022mutr3d} & \cellcolor{gray!30}0.294 & 1.498 & 0.427 & 3822 \\
		\textbf{\algname} & \cellcolor{gray!30}\textbf{0.359} & \textbf{1.320} & \textbf{0.467} & \textbf{906} \\
		\bottomrule
	\end{tabular}
	}
	\vspace{-5pt}
	\caption{\textbf{Multi-object tracking.} \algname outperforms previous end-to-end MOT techniques (with image inputs only) on all metrics. ${\text{\textdagger}}$: Tracking-by-detection method with post-association, reimplemented with BEVFormer for a fair comparison.
	}
	\label{tab:sota-track}
\end{table} 

\begin{table}[t!]
    \centering
    \scalebox{0.8}{
	\begin{tabular}{l|cccc}
		\toprule
		Method & \cellcolor{gray!30}Lanes$\uparrow$ & Drivable$\uparrow$ & Divider$\uparrow$ & Crossing$\uparrow$ \\
		\midrule
		VPN~\cite{pan2020vpn} & \cellcolor{gray!30}18.0 & 76.0 & - & - \\
		LSS~\cite{philion2020lss} & \cellcolor{gray!30}18.3 &  73.9 & - & - \\
		BEVFormer~\cite{li2022bevformer} & \cellcolor{gray!30}23.9  & \textbf{77.5} & - & - \\
        BEVerse$^{\text{\textdagger}}$~\cite{zhang2022beverse} & \cellcolor{gray!30}- & - & \textbf{30.6} & \textbf{17.2} \\
		\textbf{\algname} & \cellcolor{gray!30}\textbf{31.3} & 69.1 & 25.7 & 13.8 \\
		\bottomrule
	\end{tabular}
	}
	\vspace{-5pt}
	\caption{\textbf{Online mapping.} \algname achieves competitive performance against state-of-the-art perception-oriented methods, with comprehensive road semantics. We report segmentation IoU (\%). ${\text{\textdagger}}$: Reimplemented with BEVFormer.
	}
	\label{tab:sota-map}
\end{table} 

\begin{table}[t!]
    \centering
    \scalebox{0.8}{
	\begin{tabular}{l|cccc}
		\toprule
		Method & \cellcolor{gray!30}minADE($m$)$\downarrow$ & minFDE($m$)$\downarrow$ & MR$\downarrow$ & EPA$\uparrow$  \\
		\midrule
		PnPNet$^{\text{\textdagger}}$~\cite{liang2020pnpnet} & \cellcolor{gray!30}1.15 & 1.95 & 0.226 & 0.222 \\
		ViP3D~\cite{gu2022vip3d} & \cellcolor{gray!30}2.05 & 2.84 & 0.246 & 0.226 \\
		Constant Pos. & \cellcolor{gray!30}5.80 & 10.27 & 0.347 & - \\
		Constant Vel. & \cellcolor{gray!30}2.13 & 4.01 & 0.318 & - \\
		\textbf{\algname} & \cellcolor{gray!30}\textbf{0.71} & \textbf{1.02} & \textbf{0.151} & \textbf{0.456} \\
		\bottomrule
	\end{tabular}
	}
	\vspace{-5pt}
	\caption{\textbf{Motion forecasting.} \algname remarkably outperforms previous vision-based end-to-end methods. We also report two settings of modeling vehicles with constant positions or velocities as comparisons. ${\text{\textdagger}}$: Reimplemented with BEVFormer.
	}
	\label{tab:sota-motion}
\end{table} 

\begin{table}[t!]
    \centering
    \scalebox{0.8}{
	\begin{tabular}{l|cccc}
		\toprule
		Method & \cellcolor{gray!30}IoU-n.$\uparrow$ & IoU-f.$\uparrow$ & VPQ-n.$\uparrow$ & VPQ-f.$\uparrow$ \\
		\midrule
		FIERY~\cite{hu2021fiery} & \cellcolor{gray!30}59.4 & 36.7 & 50.2 & 29.9 \\
		StretchBEV~\cite{akan2022stretchbev} & \cellcolor{gray!30}55.5 & 37.1 & 46.0 & 29.0 \\
		ST-P3~\cite{hu2022stp3} & \cellcolor{gray!30}- & 38.9 & - & 32.1 \\
		BEVerse$^{\text{\textdagger}}$~\cite{zhang2022beverse} & \cellcolor{gray!30}61.4 & \textbf{40.9} & 54.3 & \textbf{36.1} \\
		\textbf{\algname} & \cellcolor{gray!30}\textbf{63.4} & 40.2 & \textbf{54.7} & 33.5 \\
		\bottomrule
	\end{tabular}
	}
	\vspace{-5pt}
	\caption{\textbf{Occupancy prediction.} \algname gets significant improvement in nearby areas, which are more critical for planning. ``n.'' and ``f.'' indicates near ($30\!\times\!30m$) and far ($50\!\times\!50m$) evaluation ranges respectively. ${\text{\textdagger}}$: Trained with heavy augmentations.
	}
	\label{tab:sota-occ}
\end{table}

\begin{table}[t!]
    \centering
    \scalebox{0.72}{    
	\begin{tabular}{l|cccc|cccc}
		\toprule
		\multirow{2}{*}{Method} &
		\multicolumn{4}{c|}{L2($m$)$\downarrow$} & 
		\multicolumn{4}{c}{Col. Rate(\%)$\downarrow$} \\
		& 1$s$ & 2$s$ & 3$s$ & \cellcolor{gray!30}Avg. & 1$s$ & 2$s$ & 3$s$  & \cellcolor{gray!30}Avg.\\
		\midrule
		NMP$^{\text{\textdagger}}$~\cite{zeng2019nmp} & - & - & 2.31 & \cellcolor{gray!30}- & - & - & 1.92 & \cellcolor{gray!30}- \\
		SA-NMP$^{\text{\textdagger}}$~\cite{zeng2019nmp} & - & - & 2.05 & \cellcolor{gray!30}- & - & - & 1.59 & \cellcolor{gray!30}- \\
		FF$^{\text{\textdagger}}$~\cite{hu2021safe} & 0.55 & 1.20 & 2.54 & \cellcolor{gray!30}1.43 & 0.06 & 0.17 & 1.07 & \cellcolor{gray!30}0.43 \\
		EO$^{\text{\textdagger}}$~\cite{khurana2022ssocc} & 0.67 & 1.36 & 2.78 & \cellcolor{gray!30}1.60 & 0.04 & 0.09 & 0.88 & \cellcolor{gray!30}0.33 \\
		\midrule
		ST-P3~\cite{hu2022stp3} & 1.33 & 2.11 & 2.90 & \cellcolor{gray!30}2.11 & 0.23 & 0.62 & 1.27 & \cellcolor{gray!30}0.71 \\
		\textbf{\algname} & \textbf{0.48} & \textbf{0.96} & \textbf{1.65} & \cellcolor{gray!30}\textbf{1.03} & \textbf{0.05} & \textbf{0.17} & \textbf{0.71}  & \cellcolor{gray!30}\textbf{0.31} \\
		\bottomrule
	\end{tabular}
	}
	\vspace{-5pt}
	\caption{\textbf{Planning.} \algname achieves the lowest L2 error and collision rate in all time intervals and even outperforms LiDAR-based methods (${\text{\textdagger}}$) 
 in most cases, verifying the safety of our system.
	}
	\label{tab:sota-plan}
\end{table}

\subsection{Qualitative Results}
\begin{figure*}[t!]
  \centering
  \includegraphics[width=\textwidth]{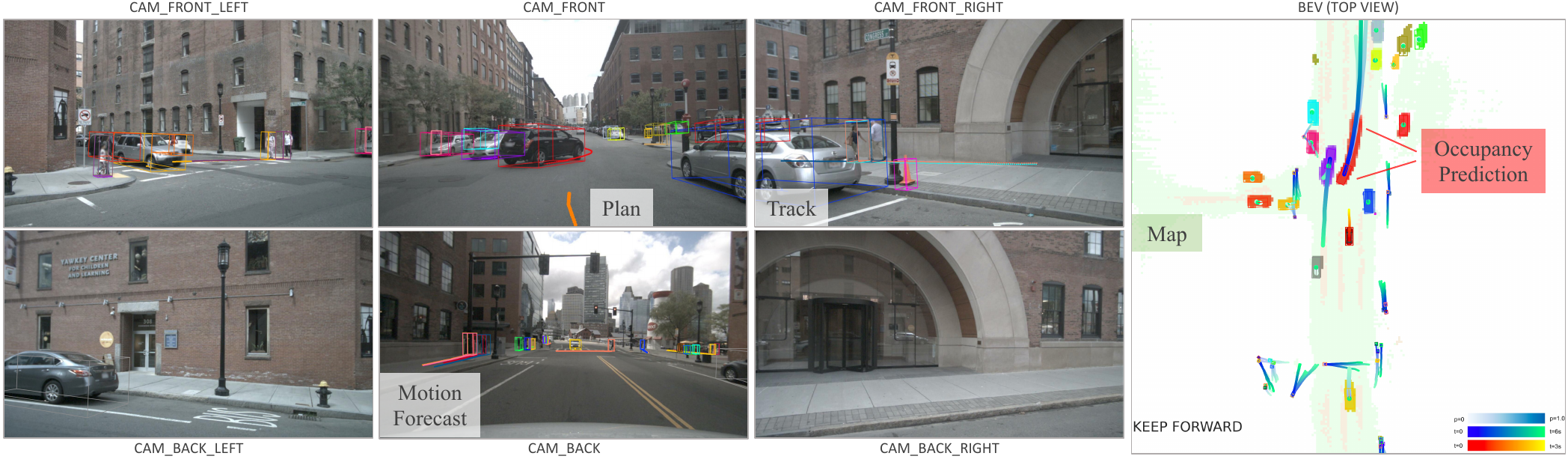}
  \vspace{-16pt}
  \caption{\textbf{Visualization results.} We show results for all tasks in surround-view images and BEV. Predictions from motion and occupancy modules are consistent, and the ego vehicle is yielding to the front black car in this case. Each agent is illustrated with a unique color. Only top-1 and top-3 trajectories from motion forecasting are selected for visualization on image-view and BEV respectively.
  }
  \label{fig:vis}
\end{figure*}

\vspace{-4pt}
\label{sec:exp-vis}
\cref{fig:vis} visualizes the results of all tasks for one complex scene. The ego vehicle drives with notice to the potential movement of a front vehicle and lane. 
In the Supplementary, we show more visualizations of challenging scenarios and one promising case for the planning-oriented design, that inaccurate results occur in prior modules while the later tasks could still recover, \eg, the planned trajectory remains reasonable though objects have a large heading angle deviation or fail to be detected in tracking results.
Besides, we analyze that failure cases of \algname are mainly under some long-tail scenarios such as large trucks and trailers, shown in the Supplementary as well.

\begin{table}[t!]
	\begin{center}
	    \centering
        \scalebox{0.64}{
			\begin{tabular}{l|cccc|cccc}
				\toprule
				
				ID & \begin{tabular}[c]{@{}c@{}}Scene-l.\\ Anch.\end{tabular} &  \begin{tabular}[c]{@{}c@{}}Goal\\ Inter.\end{tabular} & Ego $Q$ &  NLO. &  minADE$\downarrow$ & minFDE$\downarrow$ & MR$\downarrow$ & \begin{tabular}[c]{@{}c@{}}minFDE\\ -mAP$^\ast$\end{tabular}$\uparrow$  \\
				\midrule
				1 &  &  &  & & 0.844 & 1.336& 0.177& 0.246\\
				2 & \cmark  & &  &  & 0.768 & 1.159& 0.164 & 0.267\\
				3 & \cmark  & \cmark&  &  & 0.755& 1.130& 0.168 & 0.264\\
				4 & \cmark  & \cmark &\cmark & & 0.747 &1.096 & 0.156& 0.266\\
				5 & \cmark  & \cmark & \cmark & \cmark& \textbf{0.710} &\textbf{1.004} & \textbf{0.146} & \textbf{0.273}\\
				
				\bottomrule
			\end{tabular}
		}
	\end{center}
	\vspace{-15pt}
	\caption{\textbf{Ablation for designs in the motion forecasting module.} All components contribute to the ultimate performance. ``Scene-l.\,Anch.'' denotes rotated scene-level anchors. ``Goal Inter.'' means the agent-goal point interaction. ``Ego $Q$'' represents the ego-vehicle query and ``NLO.'' is the non-linear optimization strategy. $\ast$: A metric considering detection and forecasting accuracy simultaneously, and we put details in the Supplementary.}
	\label{tab:abl-traj}
\end{table}

\begin{table}[t!]
	\begin{center}
	    \centering
        \scalebox{0.72}{
			\begin{tabular}{l|ccc|cccc}
				\toprule
				
				ID & \begin{tabular}[c]{@{}c@{}}Cross.\\ Attn.\end{tabular} & \begin{tabular}[c]{@{}c@{}}Attn.\\ Mask \end{tabular} & \begin{tabular}[c]{@{}c@{}}Mask \\ Feat.\end{tabular}&  IoU-n.$\uparrow$ & IoU-f.$\uparrow$ & VPQ-n.$\uparrow$ & VPQ-f.$\uparrow$  \\
				\midrule
				1 &  &  &  & 61.2 & \textbf{39.7} & 51.5 & 31.8 \\
				2 & \cmark &  &  & 61.3 & 39.4 & 51.0 & 31.8 \\
				3 & \cmark & \cmark &  & 62.3 & \textbf{39.7} & 52.4 & 32.5 \\
				4 & \cmark & \cmark & \cmark & \textbf{62.6} & 39.5 & \textbf{53.2} & \textbf{32.8} \\
				
				\bottomrule
			\end{tabular}
		}
	\end{center}
	\vspace{-15pt}
	\caption{\textbf{Ablation for designs in the occupancy prediction module.} Cross-attention with masks and the reuse of mask feature helps improve the prediction. ``Cross. Attn.'' and ``Attn. Mask'' represent cross-attention and the attention mask in the pixel-agent interaction respectively. ``Mask Feat.'' denotes the reuse of the mask feature for instance-level occupancy. }
	\label{tab:abl-occ}
\end{table}

\begin{table}[t!]
	\begin{center}
	    \centering
        \scalebox{0.72}{
			\begin{tabular}{l|ccc|ccc|ccc}
				\toprule
				
				\multirow{2}{*}{ID} & BEV & Col. &  Occ.& \multicolumn{3}{c|}{L2$\downarrow$} &\multicolumn{3}{c}{Col. Rate$\downarrow$}  \\
				 & Att. & Loss & Optim. & 1$s$ & 2$s$ & 3$s$ & 1$s$ & 2$s$ & 3$s$ \\
				\midrule
				1 & & & & \textbf{0.44} & \textbf{0.99} & \textbf{1.71} & 0.56 & 0.88&1.64 \\
				2 & \cmark &  &  & \textbf{0.44} &1.04 & 1.81& 0.35 & 0.71 & 1.58\\
				3 & \cmark & \cmark &  & \textbf{0.44} &1.02 & 1.76& 0.30 & 0.51 & 1.39\\
				4 & \cmark & \cmark & \cmark & 0.54 & 1.09 & 1.81 & \textbf{0.13} & \textbf{0.42} & \textbf{1.05} \\
				
				\bottomrule
			\end{tabular}
		}
	\end{center}
	\vspace{-15pt}
	\caption{\textbf{Ablation for designs in the planning module.} Results demonstrate the necessity of each preceding task. ``BEV Att.'' indicates attending to BEV feature. ``Col. Loss'' denotes collision loss. ``Occ. Optim.'' is the optimization strategy with occupancy.}
	\label{tab:abl-planning}
\end{table}

\subsection{Ablation Study}
\label{sec:exp-ablation}
\vspace{-4pt}

\paragraph{Effect of designs in MotionFormer.}
\Cref{tab:abl-traj} shows that all of our proposed components described in~\cref{sec:motion-method} contribute to final performance regarding minADE, minFDE, Miss Rate and minFDE-mAP metrics.
Notably, the rotated scene-level anchor shows a significant performance boost (-15.8\%\,minADE, -11.2\%\,minFDE, +1.9 minFDE-mAP(\%)), indicating that it is essential to do motion forecasting in the scene-centric manner.
The agent-goal point interaction enhances the motion query with the planning-oriented visual feature, and surrounding agents can further benefit from considering the ego vehicle's intention.
Moreover, the non-linear optimization strategy improves the performance (-5.0\%\,minADE, -8.4\%\,minFDE, -1.0 MR(\%), +0.7 minFDE-mAP(\%)) by taking perceptual uncertainty into account in the end-to-end scenario. 

\vspace{-5pt}
\paragraph{Effect of designs in OccFormer.}
As illustrated in~\Cref{tab:abl-occ}, attending each pixel to all agents without locality constraints (Exp.2) results in slightly worse performance compared to an attention-free baseline (Exp.1). The occupancy-guided attention mask resolves the problem and brings in gain, especially for nearby areas (Exp.3, +1.0\,IoU-n.(\%), +1.4\,VPQ-n.(\%)).
Additionally, reusing the mask feature $M^t$ instead of the agent feature to acquire the occupancy feature further enhances performance.

\vspace{-5pt}
\paragraph{Effect of designs in Planner.}
We provide ablations on the proposed designs in planner in~\Cref{tab:abl-planning}, \ie, attending BEV features, training with the collision loss and the optimization strategy with occupancy.
Similar to previous research~\cite{hu2021safe,hu2022stp3}, a lower collision rate is preferred for safety over naive trajectory mimicking (L2 metric), and is reduced with all parts applied in \algname.

\section{Conclusion and Future Work}
\label{sec:conclusion}
\vspace{-4pt}

We discuss the system-level design for the autonomous driving algorithm framework.
A planning-oriented pipeline is proposed toward the ultimate pursuit for planning, namely \algname. We provide detailed analyses on the necessity of each module within perception and prediction. To unify tasks, a query-based design is proposed to connect all nodes in \algname, benefiting from richer representations for agent interaction in the environment. Extensive experiments verify the proposed method in all aspects. 

\smallskip
\noindent\textbf{Limitations and future work.} Coordinating such a comprehensive system with multiple tasks is non-trivial and needs extensive computational power, especially trained with temporal history. How to devise and curate the system for a lightweight deployment deserves future exploration. Moreover, whether or not to incorporate more tasks such as depth estimation, behavior prediction, and how to embed them into the system, are worthy future directions as well. 

\smallskip
\noindent\textbf{Acknowledgements.} This work is partially supported by National Key R\&D Program of China (2022ZD0160100), and in part by Shanghai Committee of Science and Technology (21DZ1100100), and NSFC (62206172).

{\small
\bibliographystyle{ieee_fullname}
\bibliography{egbib}
}

\appendix
\clearpage
\newpage

\noindent\textbf{\large{Appendix}}

\startcontents
{
    \hypersetup{linkcolor=black}
    \printcontents{}{1}{}
}
\newpage
\section{Task Definition}
\label{sec:task-def}

\paragraph{Detection and tracking.}
Detection and tracking are two crucial perception tasks for autonomous driving, and we focus on representing them in the 3D space to facilitate downstream usage. 3D Detection is responsible for locating surrounding objects (coordinates, length, width, height, \etc) at each time stamp; tracking aims at finding the correspondences between different objects across time stamps and associating them temporally (\ie, assigning a consistent track ID for each agent).
In the paper, we use multi-object tracking in some cases to denote the detection and tracking process.
The final output is a series of associated 3D boxes in each frame, and their corresponding features $Q_A$ are forwarded to the motion module.
Additionally, note that we have one special query named \textit{ego-vehicle query} for downstream tasks, which would not be included in the prediction-ground truth matching process and it regresses the location of ego-vehicle accordingly.

\paragraph{Online mapping.} Map intuitively embodies the geometric and semantic information of the environment, and online mapping is to segment meaningful road elements with onboard sensor data (multi-view images in our case) as a substitute for offline annotated high-definition (HD) maps.
In \algname, we model the online map into four categories: lanes, drivable area, dividers and pedestrian crossings, and we segment them in bird's-eye-view (BEV).
Similar to $Q_A$, the map queries $Q_M$ would be further utilized in the motion forecasting module to model the agent-map interaction.

\paragraph{Motion forecasting.} Bridging perception and planning, prediction plays an important role in the whole autonomous driving system to ensure final safety.
Typically, motion forecasting is an independently developed module that predicts agents' future trajectories with detected bounding boxes and HD maps. And the bounding boxes are ground truth annotations in most current motion datasets~\cite{ettinger2021waymomotion}, which is not realistic in onboard scenarios.
While in this paper, the motion forecasting module takes previously encoded sparse queries (\ie, $Q_{A}$ and $Q_{M}$) and dense BEV features $B$ as inputs, and forecasts $\mathcal{K}$ plausible trajectories in future $T$ timesteps for each agent. Besides, to be compatible with our end-to-end and scene-centric scenarios, we predict trajectories as offset according to each agent's current position.
The agent features before the last decoding MLPs, which have encoded both the historical and future information will be sent to the occupancy module for scene-level future understanding.
For the \textit{ego-vehicle query}, it predicts future ego-motion as well (actually providing a coarse planning estimation), and the feature is employed by the planner to generate the ultimate goal.

\paragraph{Occupancy prediction.} Occupancy grid map is a discretized BEV representation where each cell holds a belief indicating whether it is occupied, and the occupancy prediction task is designed to discover how the grid map changes in the future for $T_o$ timesteps with multiple agent dynamics. 
Complementary to motion forecasting which is conditioned on sparse agents, occupancy prediction is densely represented in the whole-scene level. To investigate how the scene evolves with sparse agent knowledge, our proposed occupancy module takes as inputs both the observed BEV feature $B$ and agent features $G^t$. After the multi-step agent-scene interaction (detailedly described in \cref{sec:imp-detail}), the instance-level probability map $\hat{O}_A^t\!\in\!\mathbb{R}^{N_a\!\times\!H\!\times\!W}$ is generated via matrix multiplication between occupancy feature and dense scene feature. 
To form whole-scene occupancy with agent identity preserved $\hat{O}^t\!\in\!\mathbb{R}^{H\!\times\!W}$ which is used for occupancy evaluation and downstream planning, we simply merge the instance-level probability at each timestep using pixel-wise argmax as in \cite{carion2020detr}.

\paragraph{Planning.} As an ultimate goal, the planning module takes all upstream results into consideration.
Traditional planning methods in the industry often are rule-based, formulated by ``if-else'' state machines conditioned on various scenarios which are described with prior detection and prediction results.
In our learning-based model, we take the upstream \textit{ego-vehicle query}, and the dense BEV feature $B$ as input, and predict one trajectory $\hat{\tau}$ for total $T_{p}$ timesteps. Then, the trajectory $\hat{\tau}$ is optimized with the upstream predicted future occupancy $\hat{O}$ to avoid collision and ensure final safety.

\section{The Necessity of Each Task}
\label{sec:task-necessity}

In terms of perception, \textit{tracking} in the loop as does in PnPNet~\cite{liang2020pnpnet} and ViP3D~\cite{gu2022vip3d} is proven to complement spatial-temporal features and provide history tracks for occluded agents, refraining from catastrophic decisions for downstream planning.
With the aid of HD maps~\cite{zeng2019nmp,sadat2020p3,liang2020pnpnet,gu2022vip3d} and motion forecasting, planning becomes more accurate toward higher-level intelligence. However, such information is expensive to construct and prone to be outdated, raising the demand for online \textit{mapping} without HD maps.
As for prediction, \textit{motion} forecasting~\cite{casas2018intentnet, jia2021ide, jia2022temporal, gao2020vectornet, Zhao2020tnt} generates long-term future behaviors and preserves agent identity in form of sparse waypoint outputs. 
Nonetheless, there exists the challenge to integrate non-differentiable box representation into subsequent planning module~\cite{gu2022vip3d, liang2020pnpnet}.
Some recent literature investigates another type of prediction task named \textit{occupancy}~\cite{thrun1996occgrid} prediction to assist end-to-end planning, in form of cost maps. However, the lack of agent identity and dynamics in occupancy makes it impractical to model social interactions for safe planning.
The large computational consumption of modeling multi-step dense features also leads to a much shorter temporal horizon compared to \textit{motion} forecasting. Therefore, to benefit from the two complementary types of prediction tasks for safe \textit{planning}, we incorporate both agent-centric motion and whole-scene occupancy in \algname.

\section{Related Work}
\label{sec:related-work}

\subsection{Joint perception and prediction}
Joint learning of perception and prediction is proposed to avoid the cascading error in traditional modular-independence pipelines. Similar to the motion forecasting task alone, it usually has two types of output representations: agent-level bounding boxes and scene-level occupancy grid maps.
Pioneering work FaF~\cite{luo2018faf} predicts boxes in the future and aggregates past information to produce tracklets. IntentNet~\cite{casas2018intentnet} extends it to reason about intentions and \cite{djuric2021multixnet, fadadu2022multixnetfusion} further predict future states in a refinement fashion.
Some exploit detection first and utilize agent features in the second prediction stage~\cite{casas2020spagnn, li2020contextual, peri2022futuredet}. Noticing that history information is ignored, PnPNet~\cite{liang2020pnpnet} enriches it by estimating tracking association scores to avert the non-differentiable optimization process as adopted by the tracking-by-detection paradigm~\cite{li2022uvtr, shi2022srcn3d, liu2022bevfusion, yang2022qtrack}.
Yet, all these methods rely on non-maximum suppression (NMS) in detection which still leads to information loss. ViP3D~\cite{gu2022vip3d} which is closely related to our work, employs agent queries in~\cite{zhang2022mutr3d} to forecast, taking HD map as another input.
We follow the philosophy of~\cite{zhang2022mutr3d, gu2022vip3d} in agent track queries, but also develop non-linear optimization on target trajectories to alleviate the potential inaccurate perception problem. Moreover, we introduce an ego-vehicle query for better capturing the ego behaviors in the dynamic environment, and incorporate online mapping to prevent the localization risk or high construction cost with HD map.

The alternative representation, namely the occupancy grid map, discretizes the BEV map into grid cells which holds a belief indicating if it is occupied. Wu \etal~\cite{wu2020motionnet} estimate a dense motion field, while it could not capture multimodal behaviors. Fishing Net~\cite{hendy2020fishingnet} also predicts deterministic future BEV semantic segmentation with multiple sensors.
To address this, P3~\cite{sadat2020p3} proposes non-parametric distribution of future semantic occupancy and FIERY~\cite{hu2021fiery} devises the first paradigm for multi-view cameras. A few methods improve the performance of FIERY with more sophisticated uncertainty modeling~\cite{hu2022stp3, akan2022stretchbev, zhang2022beverse}.
Notably, this representation could easily extend to motion planning for collision avoidance~\cite{sadat2020p3, casas2021mp3, hu2022stp3}, while it loses the agent identity characteristic and takes a heavy burden to computation which may constrain the prediction horizon. In contrast, we leverage agent-level information for occupancy prediction and ensure accurate and safe planning by unifying these two modes.

\subsection{Joint prediction and planning} PRECOG~\cite{rhinehart2019precog} proposes a recurrent model that conditions forecasting on the goal position of the ego vehicle, while PiP~\cite{song2020pip} generates agents' motion considering complete presumed planning trajectories. However, producing a rough future trajectory is still challenging in the real world, toward which \cite{liu2021reactive} presents a deep structured model to derive both prediction and planning from the same set of learnable costs.
\cite{ivanovic2021mats, huang2022differentiable} couple the prediction model with classic optimization methods. Meanwhile, some motion forecasting methods implicitly include the planning task by producing their future trajectories simultaneously~\cite{chai2020multipath, ngiam2021scenetransformer, kamenev2022predictionnet}.
Similarly, we encode possible behaviors of the ego vehicle in the scene-centric motion forecasting module, but the interpretable occupancy map is utilized to further optimize the plan to stay safe.

\subsection{End-to-end motion planning}
End-to-end motion planning has been an active research domain since Pomerleau~\cite{pomerleau1988alvinn} uses a single neural network that directly predicts control signals. Subsequent studies make great advances especially in closed-loop simulation with deeper networks~\cite{bojarski2016nvend}, multi-modal inputs~\cite{bansal2018chauffeurnet, codevilla2018cil, prakash2021transfuser}, multi-task learning~\cite{wu2022tcp, Chitta2022transfuserpami}, reinforcement learning~\cite{liang2018cirl, kendall2019driveaday, MaRLn, Chekroun2021gri, chen2021wor} and distillation from certain privilege knowledge~\cite{chen2020lbc, zhang2021roach, zhang2021lbw}. However, for such methods of directly generating control outputs from sensor data, the transfer from the synthetic environment to realistic application remains a problem considering their robustness and safety assurance~\cite{codevilla2019cilrs, hu2022stp3}.
Thus researchers aim at explicitly designing the intermediate representations of the network to prompt safety, where predicting how the scene evolves attracts broad interest. Some works~\cite{chitta2021neat, shao2022interfuser, hu2022mile} jointly decode planning and BEV semantic predictions to enhance interpretability, while PLOP~\cite{Buhet2020plop} adopts a polynomial formulation to provide smooth planning results for both ego vehicle and neighbors.
Cui \etal~\cite{cui2021lookout} introduce a contingency planner with diverse sets of future predictions and LAV~\cite{chen2022lav} trains the planner with all vehicles' trajectories to provide richer training data.
NMP~\cite{zeng2019nmp} and its variant~\cite{wei2021sanmp} estimate a cost volume to select the plan with minimal cost besides deterministic future perception. Though they risk producing inconsistent results between two modules, the cost map design is intuitive to recover the final plan in complex scenarios. Inspired by~\cite{zeng2019nmp}, most recent works~\cite{zeng2020dsdnet, sadat2020p3, casas2021mp3, hu2022stp3, hu2021safe} propose models that construct costs with both learned occupancy prediction and hand-crafted penalties.
However, their performances heavily rely on the tailored cost based on human experience and the distribution from where trajectories are sampled~\cite{khurana2022ssocc}.
Contrary to these approaches, we leverage the ego-motion information without sophisticated cost design and present the first attempt that incorporates the tracking module along with two genres of prediction representations simultaneously in an end-to-end model.

\section{Notations}
\label{sec:notations}
We provide a lookup table of notations and their shapes mentioned in this paper in~\Cref{tab:notation} for reference.
\begin{table*}
    \definecolor{Gray}{gray}{0.9}
	\begin{center}
			\begin{tabular}{ccl}
				\toprule
				Notation & Shape \& Params. & Description \\
				\midrule
				$Q_o$ & 900 & number of initial object queries \\
				$D$ & 256 & embed dimensions \\
				$B$ & $200 \times 200 \times 256$ &BEV feature encoded by a multi-view framework \\
				$N$ & 6 &number of transformer decoder layers for TrackFormer \\
				$N$ & 6 &number of transformer decoder layers for MapFormer \\
				$N$ & 4 &number of mask decoder layers for MapFormer \\
				$N$ & 3 &number of transformer decoder layers for MotionFormer \\
				$N$ & 5 &number of transformer decoder layers for OccFormer \\
				$N$ & 3 &number of transformer decoder layers for Planner \\
				$N_a$ & \textit{dynamic} &number of agents from TrackFormer \\
				$N_m$ & 300 &number of map queries from MapFormer \\
				$Q_A$ & $N_{a} \times 256$ &agent features from TrackFormer \\
				$P_A$ & $N_{a} \times 256$ &agent positions from TrackFormer \\
				$Q_M$ & $N_{m} \times 256$ &map features from MapFormer \\
				$\mathcal{K}$ & 6 & number of forecasting modality in MotionFormer \\
				$\tilde{\mathbf{x}}$ & $T \times 2$ & ground truth for one agent's motion forecasting \\
				$\hat{\mathbf{x}}$ & $N_{a} \times T \times 2$& prediction of motion forecasting \\
				$T$ & 12 &length of prediction timestamps in MotionFormer \\
				$Q_{\text{pos}}$ & $N_{a} \times \mathcal{K} \times 256$ & query position in MotionFormer \\
				$Q_{\text{ctx}}$ & $N_{a} \times \mathcal{K} \times 256$ &query context in MotionFormer \\
				$Q_{a}$ & $N_{a} \times \mathcal{K} \times 256$ &motion query after agent-agent interaction in MotionFormer \\
				$Q_{m}$ & $N_{a} \times \mathcal{K} \times 256$ &motion query after agent-map interaction in MotionFormer \\
				$Q_{g}$ & $N_{a} \times \mathcal{K} \times 256$ &motion query after agent-goal point interaction in MotionFormer \\
				$l$ & - &index of decoder layer \\
				$PE$ & - &sinusoidal position encoding function \\
				$I^s$ & $\mathcal{K} \times T \times 2 $ &scene-level anchor position in MotionFormer \\
				$I^a$ & $\mathcal{K} \times T \times 2 $ &agent-level anchor position in MotionFormer\\
				$\Phi$ & - &kinematic cost function set \\
				$T_o$ & 5 &length of prediction timestamps in OccFormer \\
				$G^t$ & $N_a \times 256$ & agent feature input\\
				$F^t$ &  $200 \times 200 \times 256$ & future state output \\
				$Q_X$ & $N_a \times 256$ & motion query (max-pooled on modality level) from the last layer of MotionFormer \\
				$F_{\text{ds}}^t$ & $25\times 25 \times 256$ & downscaled dense feature\\
				$F_{\text{dec}}^t$ & $200\times 200 \times 256$ &decoded dense feature after convolutional decoder\\
				$D_{\text{ds}}^t$ & $25\times 25 \times 256$ &agent-aware dense feature after pixel-agent interaction\\
				$\hat{O}_A^t$ &$N_{a} \times 200\times 200$ &instance-level probability map\\
				$\hat{O}^t$ &$200\times 200$&classical instance-agnostic occupancy map merged from $\hat{O}_A^t$ for planning\\
				$O_m^t$ & $200\times 200$ &attention mask for pixel-agent interaction\\
				$M^t$ & $N_{a} \times 256$ &mask feature\\
				$U^t$ & $N_{a} \times 256$ &occupancy feature\\
    		$T_p$ & 6 &length of planning timestamps in Planner \\
				$\hat{\tau}$ & $T_p \times 2$ &planned trajectory before the optimization with occupancy prediction \\
				$\tau^*$ & $T_p \times 2 $&ultimate plan output \\
				$\lambda$ & - &hyperparameters in cost functions, target functions, \etc \\
				\bottomrule

			\end{tabular}
	\end{center}
	\vspace{8pt}
	\caption{\textbf{Lookup table of notations and hyperparameters} in the paper. The superscript $t$ in certain notations denotes the $t^{th}$ block of OccFormer, and is omitted in descriptions for simplicity.}
	\label{tab:notation}
\end{table*} 

\section{Implementation Details}
\label{sec:imp-detail}

\subsection{Detection and Tracking}
\label{sec:detail-track}

We inherit most of the detection designs from BEVFormer~\cite{li2022bevformer} which takes a BEV encoder to transform image features into BEV feature $B$ and adopts a Deformable DETR head~\cite{zhu2020deformabledetr} to perform detection on $B$. To further conduct end-to-end tracking without heavy post association, we introduce another group of queries named track queries as in MOTR~\cite{zeng2021motr} which continuously tracks previously observed instances according to its assigned track ID. We introduce the tracking process in detail below.

\textbf{Training stage:}
At the beginning (\ie, first frame) of each training sequence, all queries are considered detection queries and predict all newborn objects, which is actually the same as BEVFormer.
Detection queries are matched to the ground truth by the Hungarian algorithm~\cite{carion2020detr}. They will be stored and updated via the query interaction module (QIM) for the next timestamp serving as track queries following MOTR~\cite{zeng2021motr}.
In the next timestamp, track queries will be directly matched with a part of ground-truth objects according to the corresponding track ID, and detection queries will be matched with the remaining ground-truth objects (newborn objects). To stabilize training, we adopt the 3D IoU metric to filter the matched queries. Only those predictions having the 3D IoU with ground-truth boxes larger than a certain threshold (0.5 in practice) will be stored and updated.

\textbf{Inference stage:}
Different from the training stage, each frame of a sequence is sent to the network sequentially, meaning that track queries could exist for a longer horizon than the training time.
Another difference emerging in the inference stage is about query updating, that we use classification scores to filter the queries (0.4 for detection queries and 0.35 for track queries in practice) instead of the 3D IoU metric since the ground truth is not available.
Besides, to avoid the interruption of tracklets caused by short-time occlusion, we use a lifecycle mechanism for the tracklets in the inference stage. Specifically, for each track query, it will be considered to disappear completely and be removed only when its corresponding classification score is smaller than 0.35 for a continuous period (2$s$ in practice).

\subsection{Online Mapping}
\label{sec:detail-map}
Following~\cite{li2022panopticseg}, we decompose the map query set into thing queries and stuff queries. The thing queries model instance-wise map elements (\ie, lanes, boundaries, and pedestrian crossings) and are matched with ground truth via bipartite matching, while the stuff query is only in charge of semantic elements (\ie, drivable area) and is processed with a class-fixed assignment. 
We set the total number of thing queries to 300 and only 1 stuff query for the drivable area. Also, we stack 6 location decoder layers and 4 mask decoder layers (we follow the structure of those layers as in~\cite{li2022panopticseg}). We empirically choose thing queries after the location decoder as our map queries $Q_M$ for downstream tasks.

\begin{figure}[t!]
  \centering
  \includegraphics[width=\linewidth]{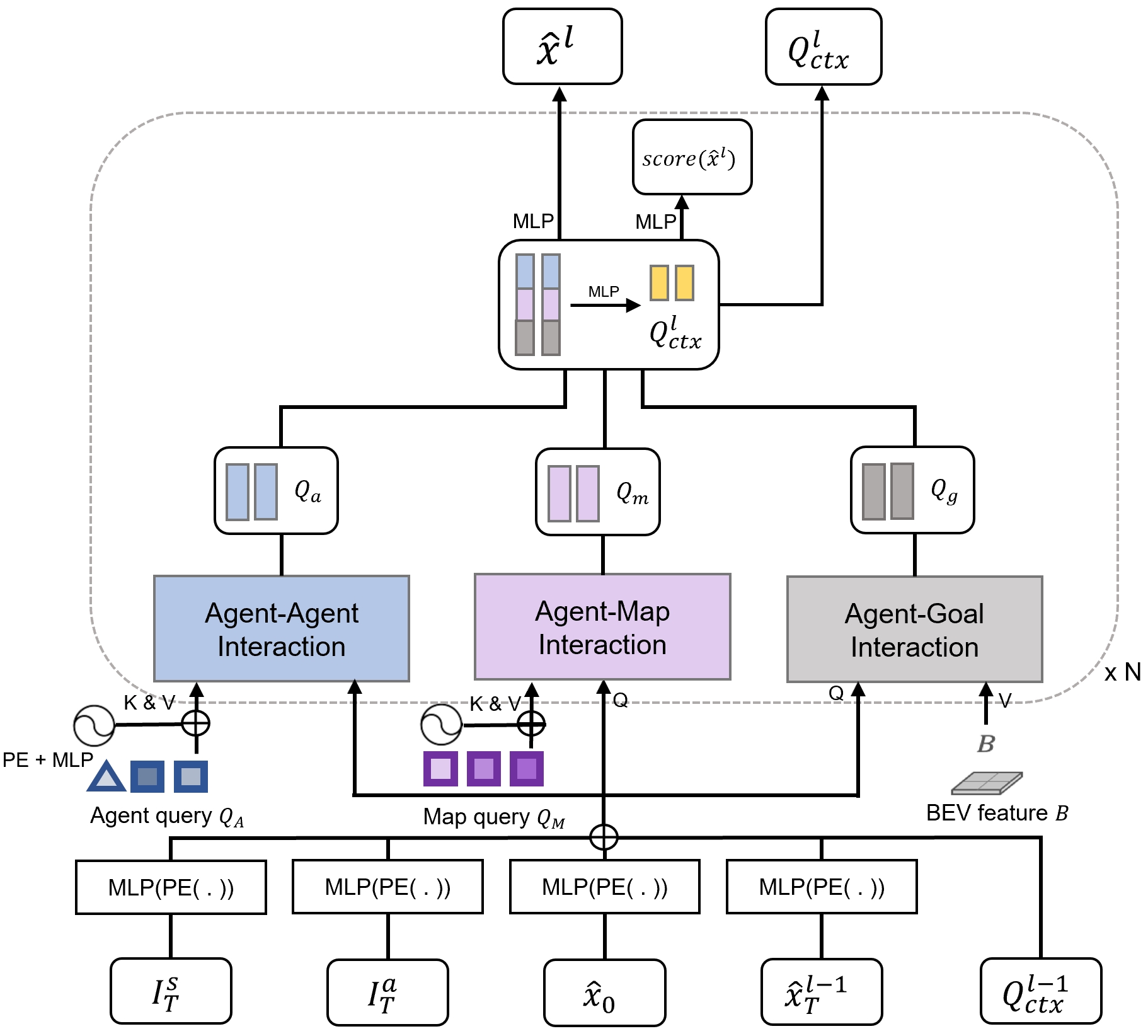}
  \caption{\textbf{MotionFormer.} It consists of $N$ stacked agent-agent, agent-map, and agent-goal interaction transformers. The agent-agent, and agent-map interaction modules are built with standard transformer decoder layers.
  The agent-goal interaction module is constructed upon the deformable cross-attention module~\cite{zhu2020deformabledetr}. $I^{s}_{T}$: the end point of scene-level anchor, $I^{a}_{T}$: the end point of clustered agent-level anchor, $\hat{x}_0$: the agent's current position, $\hat{x}_{T}^{l-1}$: the predicted goal point from the previous layer, $Q_{\text{ctx}}^{l-1}$: query context from the previous layer.
  }
  \label{fig:motion}
\end{figure}

\begin{figure}[t!]
  \centering
  \scalebox{0.85}{
  \includegraphics[width=\linewidth]{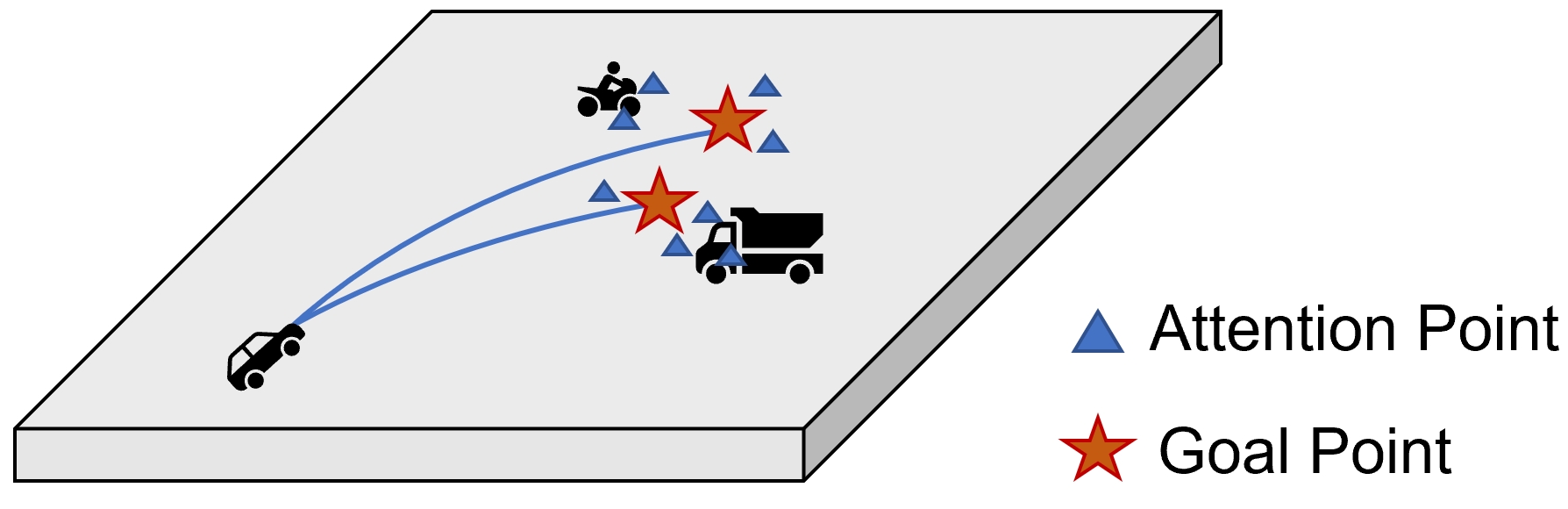}
  }
  \caption{\textbf{Illustration of agent-goal interaction Module.} The BEV visual feature is sampled near each agent's goal points with deformable cross-attention.
  }
  \label{fig:goal-interaction}
\end{figure}

\subsection{Motion Forecasting}
\label{sec:detail-motion}
To better illustrate the details, we provide a diagram as shown in~\cref{fig:motion}. 
Our MotionFormer takes $I^{a}_{T}$, $I^{s}_{T}$, $\hat{x} _0$, $\hat{x}_{T}^{l-1} \in \mathbb{R}^{\mathcal{K}\times 2}$ to embed query position, and takes $Q_{ctx}^{l-1}$ as query context. Specifically, the anchors are clustered among training data of all agents by the k-means algorithm, and we set $\mathcal{K}\!=\!6$ which is compatible with our output modalities.
To embed the scene-level prior, the anchor $I^{a}_{T}$ is rotated and translated into the global coordinate frame according to each agent's current location and heading angle, which is denoted as $I^{s}_{T} $, as shown in~\cref{eq:sl-anchors},
\begin{equation}
\label{eq:sl-anchors}
    I_{i, T}^{s} = R_{i} I_{T}^{a} + T_{i},
\end{equation}
where $i$ is the index of the agent, and it is omitted later for brevity. To facilitate the coarse-to-fine paradigm, we also adopt the goal point predicted from the previous layer $\hat{x}_{T}^{l-1} $. In the meantime, the agent's current position is broadcast across the modality, denoted as $\hat{x}_0 $.
Then, MLPs and sinusoidal positional embeddings are applied for each of the prior positional knowledge and we summarize them as the query position $Q_{\text{pos}} \in \mathbb{R}^{\mathcal{K}\times \mathcal{D}}$, which is of the same shape as the query context $Q_{ctx}$. $Q_{\text{pos}}$ and $Q_{\text{ctx}}$ together build up our motion query. We set $\mathcal{D}$ to 256 throughout MotionFormer.

As shown in~\cref{fig:motion}, our MotionFormer consists of three major transformer blocks, \ie, agent-agent, agent-map and agent-goal interaction modules.
The agent-agent, agent-map interaction modules are built with standard transformer decoder layers, which are composed of a multi-head self-attention (MHSA) layer and a multi-head cross-attention (MHCA) layer, a feed-forward network (FFN) and several residual and normalization layers in between~\cite{carion2020detr}.
Apart from the agent queries $Q_{A}$ and map queries $Q_M$, we also add the positional embeddings to those queries with sinusoidal positional embedding followed by MLP layers.
The agent-goal interaction module is built upon deformable cross-attention module~\cite{zhu2020deformabledetr}, where the goal point from the previously predicted trajectory ($R_i\hat{x}_{i,T}^{l-1} + T_{i} $) is adopted as the reference point, as shown in~\cref{fig:goal-interaction}.
Specifically, we set the number of sampled points to 4 per trajectory, and 6 trajectories per agent as we mention above.
The output features of each interaction module are concatenated and projected with MLP layers to dimension $\mathcal{D}\!=\!256$. Then, we use Gaussian Mixture Model to build each agent's trajectories, where $\hat{x}_{l} \in \mathcal{R}^{\mathcal{K} \times T \times 5}$. We set the prediction time horizon $T$ to 12 (6 seconds) in \algname.
Note that we only take the first two of the last dimension (\ie, $x$ and $y$) as final output trajectories. Besides, the scores of each modality are also predicted ($score(\hat{x}_{l}) \in \mathcal{R}^{\mathcal{K}}$). We stack the overall modules for $N$ times, and $N$ is set to 3 in practice.

\subsection{Occupancy Prediction} 
\label{sec:detail-occ}

\begin{figure}[t!]
  \centering
  \includegraphics[width=0.9\linewidth]{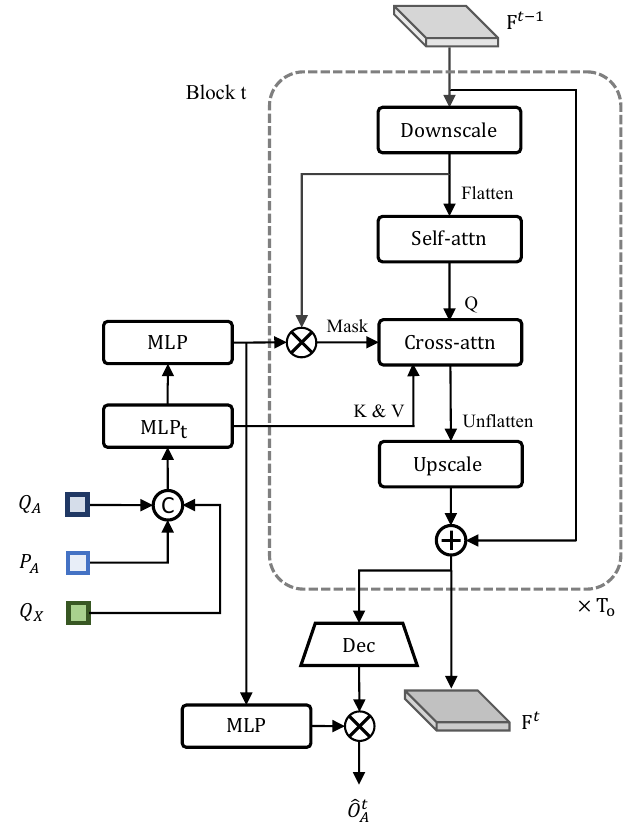}
  \vspace{-8pt}
  \caption{\textbf{OccFormer.} It is composed of $T_o$ sequential blocks where $T_o$ is the temporal horizon (including current and future frames) and each block is responsible for generating occupancy of one specific frame. 
  We incorporate both dense scene features and sparse agent features, which are encoded from upstream track query $Q_A$, agent position $P_A$ and motion query $Q_X$, to inject agent-level knowledge into future scene representations.
  We form instance-level occupancy $\hat{O}_A^t$ via a matrix multiplication between agent-level feature and decoded dense feature at the end of each block.
  }
  \label{fig:occ}
\end{figure}

Given the BEV feature from upstream modules, we first downsample it by /4 with convolutional layers
for efficient multi-step prediction, then pass it to our proposed OccFormer. OccFormer is composed of $T_o$ sequential blocks shown in~\cref{fig:occ}, where $T_o\!=\!5$ is the temporal horizon (including current and future frames) and each block is responsible for generating occupancy of one specific frame.
Different from prior works which are short of agent-level knowledge, our proposed method incorporates both dense scene features and sparse agent features when unrolling the future representations.
The dense scene feature is from the output of the last block (or the observed feature for current frame) and it's further downscaled (/8) by a convolution layer to reduce computation for pixel-agent interaction.
The sparse agent feature is derived from the concatenation of track query $Q_A$, agent positions $P_A$, and motion query $Q_X$, and it is then passed to a temporal-specific MLP for temporal sensitivity.
We conduct pixel-level self-attention to model the long-term dependency required in some rapidly changing scenes, then perform scene-agent incorporation by attending each pixel of the scene to corresponding agents.
To enhance the location alignment between agents and pixels, we restrict the cross-attention with an attention mask which is generated by a matrix multiplication between mask feature and downscaled scene feature, where the mask feature is produced by encoding agent feature with an MLP.
We then upsample the attended dense feature to the same resolution as input $F^{t-1}$ (/4) and add it with $F^{t-1}$ as a residual connection for stability.
The resulting feature $F^{t}$ is both sent to the next block and a convolutional decoder for predicting occupancy at the original BEV resolution (/1). We reuse the mask feature and pass it to another MLP to form occupancy feature, and the instance-level occupancy is therefore generated by a matrix multiplication between occupancy feature and decoded dense feature $F^t_{\text{dec}}$ (/1). 
Note that the MLP layer for mask feature, the MLP layer for occupancy feature, and the convolutional decoder are shared across all $T_o$ blocks while other components are independent in each block. Dimensions of all dense features and agent features are 256 in OccFormer.

\subsection{Planning}
\label{sec:detail-plan}
\begin{figure}[t!]
  \centering
  \includegraphics[width=0.8\linewidth]{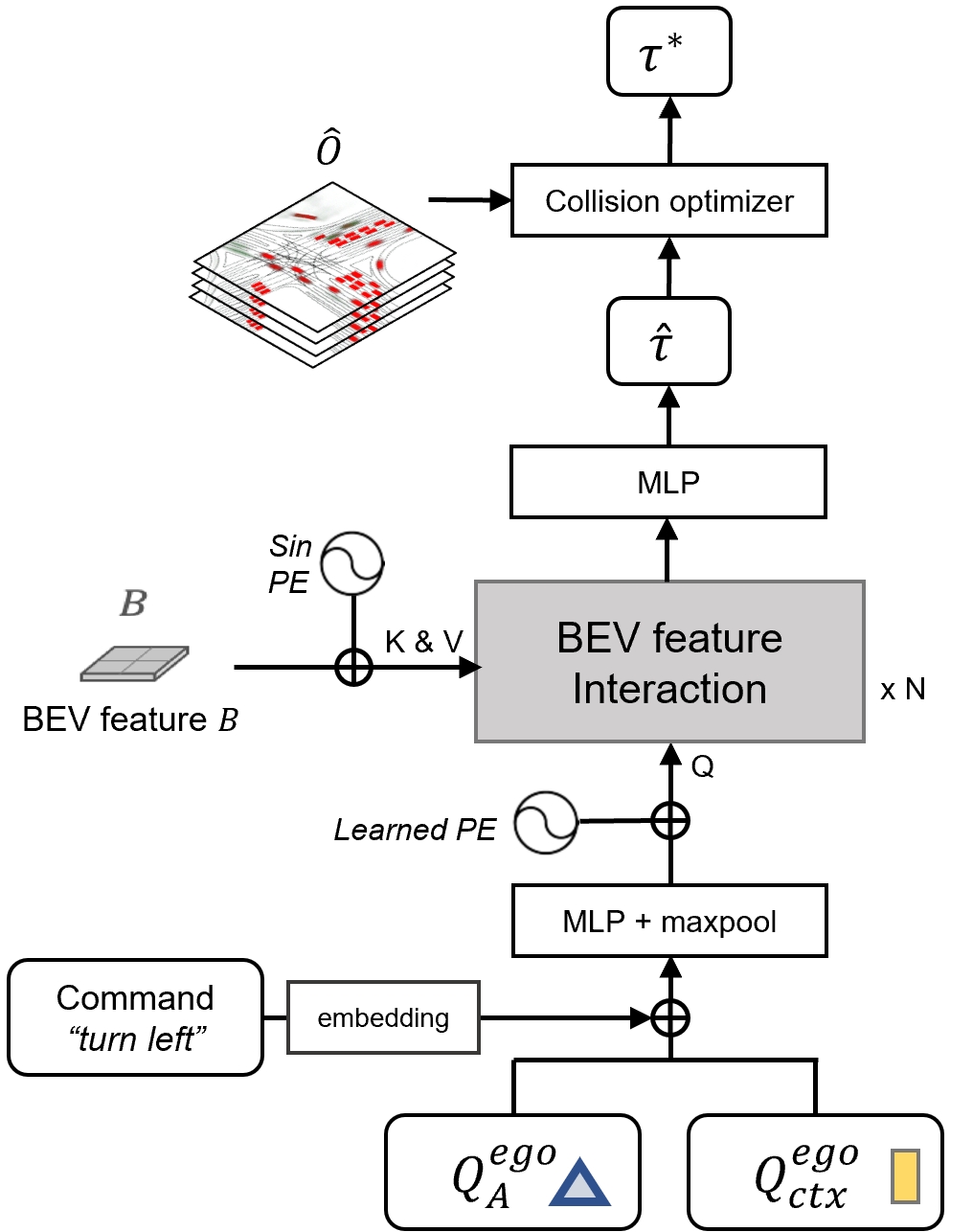}
  \caption{\textbf{Planner.}
  $Q_A^{\text{ego}}$ and $Q_{\text{ctx}}^{\text{ego}}$ are the \textit{ego-vehicle query} from the tracking and motion forecasting modules, respectively.
  Along with the high-level command, they are encoded with MLP layers followed by a max-pooling layer across the modality dimension, where the most salient modal features are selected and aggregated.
  The BEV feature interaction module is built with standard transformer decoder layers, and it is stacked for $N$ layers. 
  }
  \label{fig:plan}
\end{figure}

As shown in~\cref{fig:plan}, our planner takes the ego-vehicle query generated from the tracking and motion forecasting module, which is symbolized with the blue triangle and yellow rectangle respectively.
These two queries, along with the command embedding, are encoded with MLP layers followed by a max-pooling layer across the modality dimension, where the most salient modal features are selected and aggregated.
The BEV feature interaction module is built with standard transformer decoder layers, and it is stacked for $N$ layers, where we set $N\!=\!3$ here. Specifically, it cross-attends the dense BEV feature with the aggregated plan query.
More qualitative results can be found in~\cref{sec:exp-qualitative} showing the effectiveness of this module. To embed location information, we fuse the plan-query with learned position embedding and the BEV feature with sinusoidal positional embedding.
We then regress the planning trajectory with MLP layers, which is denoted as $\hat{\tau} \in \mathcal{R}^{T_{p} \times 2}$. Here we set $T_{p}\!=
\!6$ (3 seconds). Finally, we devise the collision optimizer for obstacle avoidance, which takes the predicted occupancy $\hat{O}$ and trajectory $\hat{\tau}$ as input as \cref{eq:col-cost} in the main paper. We set $d\!=\!5$, $\sigma\!=\!1.0$, $\lambda_{\text{coord}}\!=\!1.0$, $\lambda_{\text{obs}}\!=\!5.0$.

\subsection{Training Details}
\label{sec:detail-train}

\paragraph{Joint learning.}
\algname is trained in two stages which we find more stable.
In \textbf{stage one}, we pre-train perception tasks including tracking and online mapping to stabilize perception predictions.
Specifically, we load corresponding pre-trained BEVFormer~\cite{li2022bevformer} weights to \algname for fast convergence including image backbone,  FPN, BEV encoder and detection decoder except for object query embeddings (due to the additional \textit{ego-vehicle query}). We stop the gradient back-propagation in the image backbone to reduce memory cost and train \algname for 6 epochs with tracking and online mapping losses as follows:
\begin{equation}
\label{eq:all-loss-1}
    L_1 = L_{\text{track}} + L_{\text{map}}.
\end{equation}
In \textbf{stage two}, we keep the image backbone frozen as well, and additionally freeze BEV encoder, which is used for view transformation from image view to BEV, to further reduce memory consumption with more downstream modules. \algname now is trained with all task losses including tracking, mapping, motion forecasting, occupancy prediction, and planning for 20 epochs (for various ablation studies in main paper, it's trained for 8 epochs for efficiency):
\begin{equation}
\label{eq:all-loss}
    L_2 = L_{\text{track}} + L_{\text{map}} + L_{\text{motion}} + L_{\text{occ}} + L_{\text{plan}}.
\end{equation}
Detailed losses and hyperparameters within each term of $L_1$ and $L_2$ are described below individually. The length of each training sequence (at each step) for tracking and BEV feature aggregation\cite{li2022bevformer} in both stages is 5 (3 in ablation studies for efficiency).

\paragraph{Detection\&tracking loss.} Following BEVFormer~\cite{li2022bevformer}, the \textit{Hungarian loss} is adopted for each paired result, which is a linear combination of a Focal loss~\cite{lin2017focal} for class labels and an $l_1$ for 3D boxes localization.
In terms of the matching strategy, candidates from newborn queries are paired with ground truth objects through bipartite matching, and predictions from track queries inherit the assigned ground truth index from previous frames.
Specifically, $L_{\text{track}} = \lambda_{\text{focal}} L_{\text{focal}} + \lambda_{l_1} L_{l_1}$, where $\lambda_{\text{focal}}\!=\!2$ and $\lambda_{l_1}\!=\!0.25$.

\paragraph{Online mapping loss.} As in~\cite{li2022panopticseg}, this includes thing losses for lanes, dividers, and contours, also a stuff loss for the drivable area, where Focal loss is responsible for classification, L1 loss is responsible for thing bounding boxes, Dice loss and GIoU loss\cite{rezatofighi2019giou} account for segmentation.
Detailedly, $L_{\text{map}} = \lambda_{\text{focal}} L_{\text{focal}} + \lambda_{l_1} L_{l_1} + \lambda_{\text{giou}} L_{\text{giou}} + \lambda_{\text{dice}} L_{\text{dice}}$, with $\lambda_{\text{focal}}\!=\!\lambda_{\text{giou}}\!=\!\lambda_{\text{dice}}\!=\!2$ and $\lambda_{l_1}\!=\!0.25$.

\paragraph{Motion forecasting loss.} Like most of the prior methods, we model the multimodal trajectories as gaussian mixtures, and use the multi-path loss~\cite{chai2020multipath, varadarajan2021multipath++}, which includes a classification score loss $L_{\text{cls}}$ and a negative log-likelihood loss term $L_{\text{nll}}$, and $\lambda$ denotes the corresponding weight:
$L_{\text{motion}} = \lambda_{\text{cls}} L_{\text{cls}} + \lambda_{\text{reg}} L_{\text{nll}}$, where $\lambda_{\text{cls}}\!\!=\!\!\lambda_{\text{reg}}\!\!=\!\!0.5$.
To ensure the temporal smoothness of trajectories, we predict agents' speed at each timestep first and accumulate it across time to obtain their final trajectories~\cite{jia2022temporal}.

\paragraph{Occupancy prediction loss.} The output of instance-level occupancy prediction is a binary segmentation of each agent, therefore we adopt binary cross-entropy and Dice loss~\cite{milletari2016dice} as the occupancy loss.
Formally, $L_{\text{occ}} = \lambda_{\text{bce}} L_{\text{bce}} + \lambda_{\text{dice}} L_{\text{dice}}$, with $\lambda_{\text{bce}}\!=\!5$ and $\lambda_{\text{dice}}\!=\!1$ here.
Additionally, since the attention mask in the pixel-agent interaction module could be seen as a coarse prediction, we employ an auxiliary occupancy loss with the same form to supervise it.

\paragraph{Planning loss.} Safety is the most crucial factor in planning. Therefore, apart from the naive imitation $l_2$ loss, we employ another collision loss which keeps the planned trajectory away from obstacles as follows:
\begin{equation}
\label{eq:col-iouloss}
    L_{\text{col}}(\hat{\tau}, \delta) = \sum_{i, t} \texttt{IoU}(box(\hat{\tau}_{t}, w+\delta, l+\delta), b_{i, t})),   
\end{equation}
\begin{equation}
\label{eq:plan-loss}
    L_{\text{plan}} = \lambda_{\text{imi}} |\hat{\tau}, \tilde{\tau}|_2 + \lambda_{\text{col}} \sum_{(\omega,\delta) } \omega  L_{\text{col}}(\hat{\tau}, \delta),
\end{equation}
where $\lambda_{\text{imi}}\!=\!1$, $\lambda_{\text{col}}\!=\!2.5$, $(\omega,\delta)$ is a weight-value pair considering additional safety distance, $box(\hat{\tau}_{t}, w\!+\!\delta, l\!+\!\delta)$
represents the ego bounding box with an increased size at timestamp $t$ to keep a larger safe distance, and $b_{i,t}$ indicates each agent forecasted in the scene. In practice, we set $(\omega,\delta)$ to $(1.0, 0.0), (0.4, 0.5), (0.1, 1.0)$.

\section{Experiments}
\label{sec:exp}

\subsection{Protocols}
\label{sec:exp-protocol}

We follow most of the basic training settings as in BEVFormer~\cite{li2022bevformer} for both two stages with a batch size of 1, a learning rate of $2\!\times\!10^{-4}$, learning rate multiplier of the backbone 0.1 and AdamW optimizer~\cite{adamw} with a weight decay of $1\!\times\!10^{-2}$.
The default size of BEV size is $200\!\times\!200$, covering BEV ranges of [-51.2$m$, 51.2$m$] for both X and Y axis with the interval as 0.512$m$.
More hyperparameters related to feature dimensions are shown in~\Cref{tab:notation}. Experiments are conducted with 16 NVIDIA Tesla A100 GPUs.

\subsection{Metrics}
\label{sec:exp-metric}
\paragraph{Multi-object tracking.}
Following the standard evaluation protocols, we use \textbf{AMOTA} (Average Multi-object Tracking Accuracy), \textbf{AMOTP} (Average Multi-object Tracking Precision), \textbf{Recall}, and \textbf{IDS} (Identity Switches) to evaluate the 3D tracking performance of \algname on nuScenes dataset.
AMOTA and AMOTP are computed by integrating MOTA (Multi-object Tracking Accuracy) and MOTP (Multi-object Tracking Precision) values over all recalls:
\begin{equation}
    \label{eq:amota}
    \text{AMOTA} = \frac{1}{n-1}\sum_{r\in\{\frac{1}{n-1}, \frac{2}{n-1},...,1\}}
    \text{MOTA}_{r},
\end{equation}
\begin{equation}
    \label{eq:mota}
    \text{MOTA}_{r} = \mathop{\max}(0, 1-\frac{\text{FP}_{r}+\text{FN}_{r}+\text{IDS}_{r}-(1-r)\text{GT}}{r\text{GT}}),
\end{equation}
where $\text{FP}_{r}$, $\text{FN}_{r}$, and $\text{IDS}_{r}$ represent the number of false positives, false negatives and identity switches computed at the corresponding recall $r$, respectively. GT stands for the number of ground truth objects in this frame. AMOTP can be defined as:
\begin{equation}
    \label{eq:amotp}
    \text{AMOTP} =
    \frac{1}{n-1}\sum_{r\in\{\frac{1}{n-1}, \frac{2}{n-1},...,1\}}
    \frac{\sum_{i,t}d_{i,t}}{\text{TP}_r},
\end{equation}
where $d_{i,t}$ denotes the position error (in $x$ and $y$ axis) of matched track $i$ at time stamp $t$, and $\text{TP}_r$ is the number of true positives at the corresponding recall $r$.

\paragraph{Online mapping.} We have four categories for the online mapping task, \ie, lanes, boundaries, pedestrian crossings and drivable area. We calculate the intersection-over-union (\textbf{IoU}) metric for each class between the network outputs and ground truth maps.

\paragraph{Motion forecasting.} On one hand, following the standard motion prediction protocols, we adopt conventional metrics, including \textbf{minADE} (minimum Average Displacement Error), \textbf{minFDE} (minimum Final Displacement Error) and \textbf{MR} (Miss Rate). Similar to the prior works~\cite{luo2018faf, liang2020pnpnet, peri2022futuredet}, these metrics are only calculated within matched TPs, and we set the matching threshold to 1.0$m$ in all of our experiments.
As for the MR, we set the miss FDE threshold to 2.0$m$. On the other hand, we also employ recently proposed end-to-end metrics, \ie, \textbf{EPA} (End-to-end Prediction Accuracy)~\cite{gu2022vip3d} and \textbf{minFDE-AP}~\cite{peri2022futuredet}.
For EPA, we use the same setting as in ViP3D~\cite{gu2022vip3d} for a fair comparison. For minFDE-AP, we do not separate ground truth into multiple bins (static, linear, and non-linearly moving sub-categories) for simplicity.
Specifically, only when an object's perception location and its min-FDE are within the distance threshold (1.0$m$ and 2.0$m$ respectively), it would be counted as a TP for the AP (average precision) calculation. 
Similarly to the prior works, we merge the car, truck, construction vehicle, bus, trailer, motorcycle, and bicycle as the vehicle category, and all the motion forecasting metrics provided in the experiments are evaluated on the vehicle category.

\begin{table*}[t!]
    \definecolor{Gray}{gray}{0.9}
	\begin{center}
		\resizebox{\textwidth}{!}{
			\begin{tabular}{l|c|ccc|cc|cccc|cccc|cc}
				\toprule
				\multirow{2}{*}{Methods} & 
				\multirow{2}{*}{Encoder} & 
				\multicolumn{3}{c|}{Tracking} & 
				\multicolumn{2}{c|}{Mapping} & 
				\multicolumn{4}{c|}{Motion Forecasting} & 
				\multicolumn{4}{c|}{Occupancy Prediction} & \multicolumn{2}{c}{Planning} \\
				& & AMOTA$\uparrow$ & AMOTP$\downarrow$ &IDS$\downarrow$ & IoU-lane$\uparrow$ & IoU-road$\uparrow$ & minADE$\downarrow$ & minFDE$\downarrow$ & MR$\downarrow$ & EPA$\uparrow$ & IoU-n.$\uparrow$ & IoU-f.$\uparrow$ & VPQ-n.$\uparrow$ & VPQ-f.$\uparrow$ & avg.L2$\downarrow$ & avg.Col.$\downarrow$ \\
				\midrule
				\textbf{\algname}-S & R50  & 0.241 & 1.488 & 958 & 0.315 & 0.689 & 0.788 & 1.126 & 0.156 & 0.381 & 59.4 & 35.6 & 49.2 & 28.9 & 1.04 & 0.32 \\
				\textbf{\algname}-B & R101 & 0.359 & 1.320 & \textbf{906} & 0.313 & 0.691 & \textbf{0.708} & \textbf{1.025} & \textbf{0.151} & 0.456 & 63.4 & 40.2 & 54.7 & 33.5 & 1.03 & 0.31 \\
				\textbf{\algname}-L & V2-99$^\ast$  & \textbf{0.409} & \textbf{1.259} & 1583 & \textbf{0.323} & \textbf{0.709} & 0.723 & 1.067 & 0.158 & \textbf{0.508} & \textbf{64.1} & \textbf{42.6} & \textbf{55.8} & \textbf{36.9} & \textbf{1.03} & \textbf{0.29}\\
				\bottomrule
			\end{tabular}
		}
	\end{center}
	\vspace{-10pt}
	\caption{\textbf{Comparisons between three variations of UniAD.}
	$\ast$: pre-trained with extra depth data~\cite{park2021dd3d}.
        }
	\label{tab:SOTA_2}
\end{table*} 

\paragraph{Occupancy prediction.} We evaluate the quality of predicted occupancy in both whole-scene level and instance-level following~\cite{hu2021fiery, zhang2022beverse}. Specifically, The \textbf{IoU} measures the whole-scene categorical segmentations which is instance-agnostic, while the Video Panoptic Quality (\textbf{VPQ})~\cite{kim2020vpq} takes into account each instance's presence and consistency over time. The VPQ metric is calculated as follows:
\begin{equation}
    \text{VPQ} = \sum_{t=0}^H \frac{\sum_{(p_t,q_t) \in \text{TP}_t} \texttt{IoU}(p_t,q_t)}{|\text{TP}_t| + \frac{1}{2}|\text{FP}_t| + \frac{1}{2}|\text{FN}_t|},
\end{equation}
where $H$ is the future horizon and we set $H\!=\!4$ (which leads to $T_o\!=\!5$ including the current timestamp) as in~\cite{hu2021fiery, zhang2022beverse}, covering 2.0$s$ consecutive data at 2Hz. $\text{TP}_t$, $\text{FP}_t$, and $\text{FN}_t$ are the set of true positives, false positives, and false negatives at timestamp $t$ respectively. Both two metrics are evaluated under two different BEV ranges, \textbf{near} (``-n.'') for $30m\!\times\!30m$ and \textbf{far} (``-f.'') for $100m\!\times\!100m$ around the ego vehicle. We evaluate the results of the current step ($t\!=\!0$) and the future 4 steps together on both metrics.

\begin{table}[t!]
    \definecolor{Gray}{gray}{0.9}
	\begin{center}
        \scalebox{0.65}{
			\begin{tabular}{l|cccccc|ccc}
				\toprule
				ID & Det. & Track & Map & Motion & Occ. & Plan &
				\#Params & FLOPs & FPS  \\
				\midrule
				0\cite{zhang2022beverse} & \cmark &  & \cmark &  & \cmark &  & 102.5M  & 1921G & - \\
				\midrule
				1 & \cmark &  &  &  &  &  &  65.9M & 1324G & 4.2  \\
				2 & \cmark & \cmark &  &  &  &  & 68.2M & 1326G & 2.7  \\
				3 & \cmark & \cmark & \cmark &  &  &  & 95.8M & 1520G & 2.2  \\
				4 & \cmark & \cmark & \cmark & \cmark &  &  & 108.6M & 1535G & 2.1  \\
				5 & \cmark & \cmark & \cmark & \cmark & \cmark &  & 122.5M & 1701G & 2.0  \\
				6 & \cmark & \cmark & \cmark & \cmark & \cmark & \cmark & 125.0M & 1709G &  1.8 \\
				\bottomrule
			\end{tabular}
		}
	\end{center}
	\vspace{-10pt}
	\caption{\textbf{Computational complexity and runtime} with different modules incorporated. ID.1 is similar to original BEVFormer~\cite{li2022bevformer}, and ID. 0 (BEVerse-Tiny)~\cite{zhang2022beverse} is an MTL framework.}
	\label{tab:abl-module-computation}
\end{table} 

\paragraph{Planning.} We adopt the same metrics as in ST-P3~\cite{hu2022stp3}, \ie,  \textbf{L2 error} and \textbf{collision rate} at various timestamps.

\subsection{Model complexity and Computational cost}
We measure the complexity of \algname and runtime on an Nvidia Tesla A100 GPU, as depicted in~\Cref{tab:abl-module-computation}.
Though the decoder part of tasks brings a certain amount of parameters, the computational complexity mainly comes from the encoder part, compared to the original BEVFormer detector (ID. 1).
We also provide a comparison with the recent BEVerse~\cite{zhang2022beverse}. \algname owns more tasks, achieves superior performance, and has \textit{lower} FLOPs - indicating affordable budget to additional computation cost.

\subsection{Model scale}
\label{sec:exp-sota2}
We provide three variations of \algname under different model scales as shown in~\Cref{tab:SOTA_2}. The chosen image backbones for image-view feature extraction are ResNet-50\cite{he2016resnet}, ResNet-101 and VoVNet 2-99\cite{lee2019vovnet} for \algname-S, \algname-B and \algname-L respectively.
Since the model scale (image encoder) mainly influences the BEV feature quality, we could observe that the perceptual scores improve with a larger backbone, which further could lead to better prediction and planning performance.

\subsection{Qualitative results}
\label{sec:exp-qualitative}

\begin{figure*}[t!]
  \centering
  \includegraphics[width=0.6\linewidth]{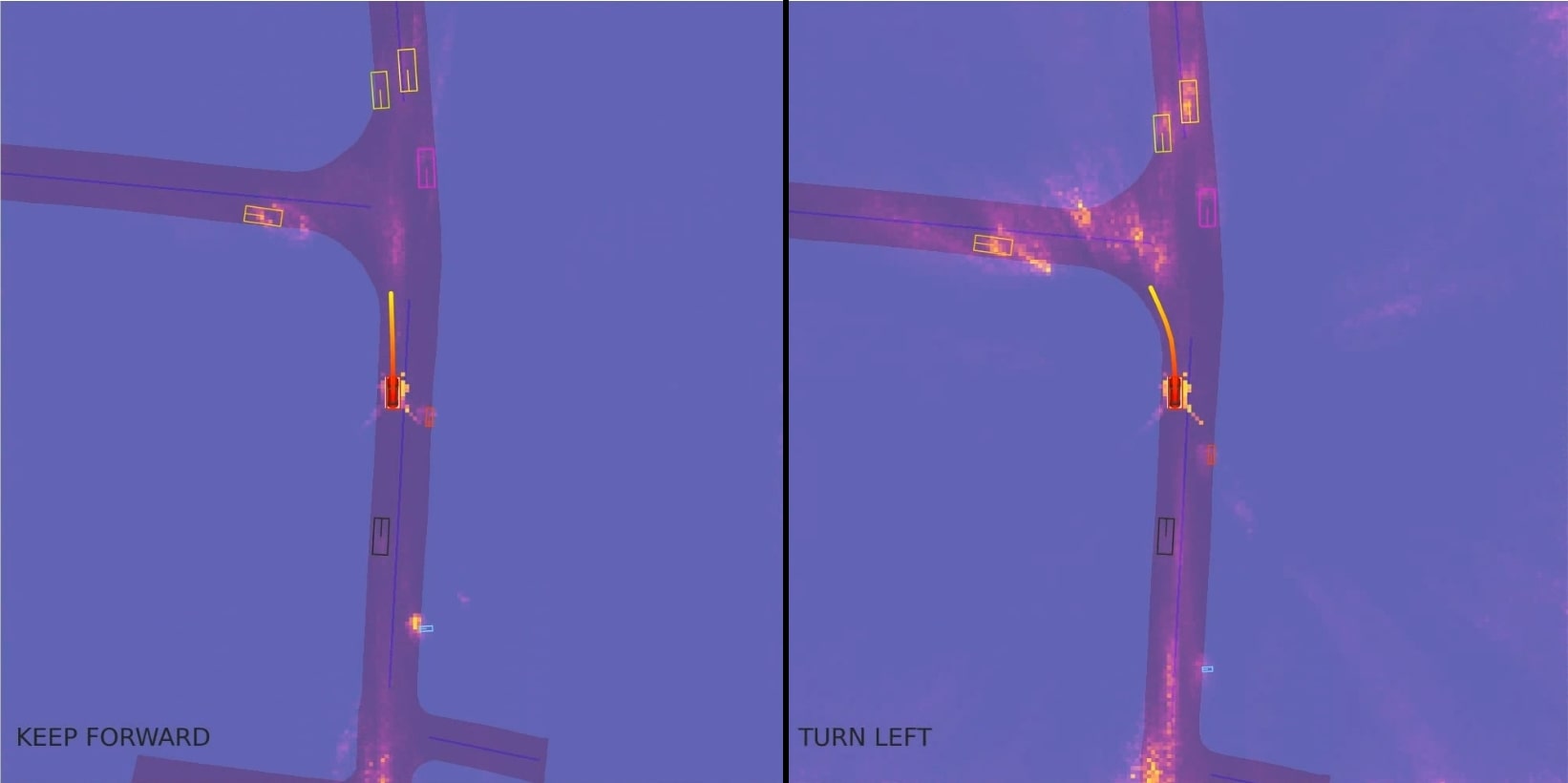}
  \caption{\textbf{Effectiveness of navigation command and attention mask visualization.} Here we demonstrate how attention is paid in accordance with the navigation command. We render the attention mask from the BEV interaction module in the planning module, the predicted tracking bounding boxes as well as the planned trajectory. The navigation command is printed on the bottom left, and the HD Map is rendered only for reference. From left to right, we show two consecutive frames in a time sequence but with different navigation commands. We can observe that the planned trajectory varies largely according to the command. Also, much attention is paid to the goal lane as well as the critical agents that are yielding to our ego vehicle.}
  \label{fig:vis-supp-attn_mask}
\end{figure*}

\paragraph{Attention mask visualization.} To investigate the internal mechanism and show its explainability, we visualize the attention mask of the cross-attention module in the planner. As shown in~\cref{fig:vis-supp-attn_mask}, the predicted tracking bounding boxes, planned trajectory, and the ground truth HD Map are rendered for reference, and the attention mask is overlayered on top.
From left to right, we show two consecutive frames in a time sequence but with different navigation commands. We can observe that the planned trajectory
varies largely according to the command. Also, much attention is paid to the goal lane as well as the critical agents that are yielding to our ego vehicle.

\begin{figure*}[t!]
  \centering
  \includegraphics[width=0.98\textwidth]{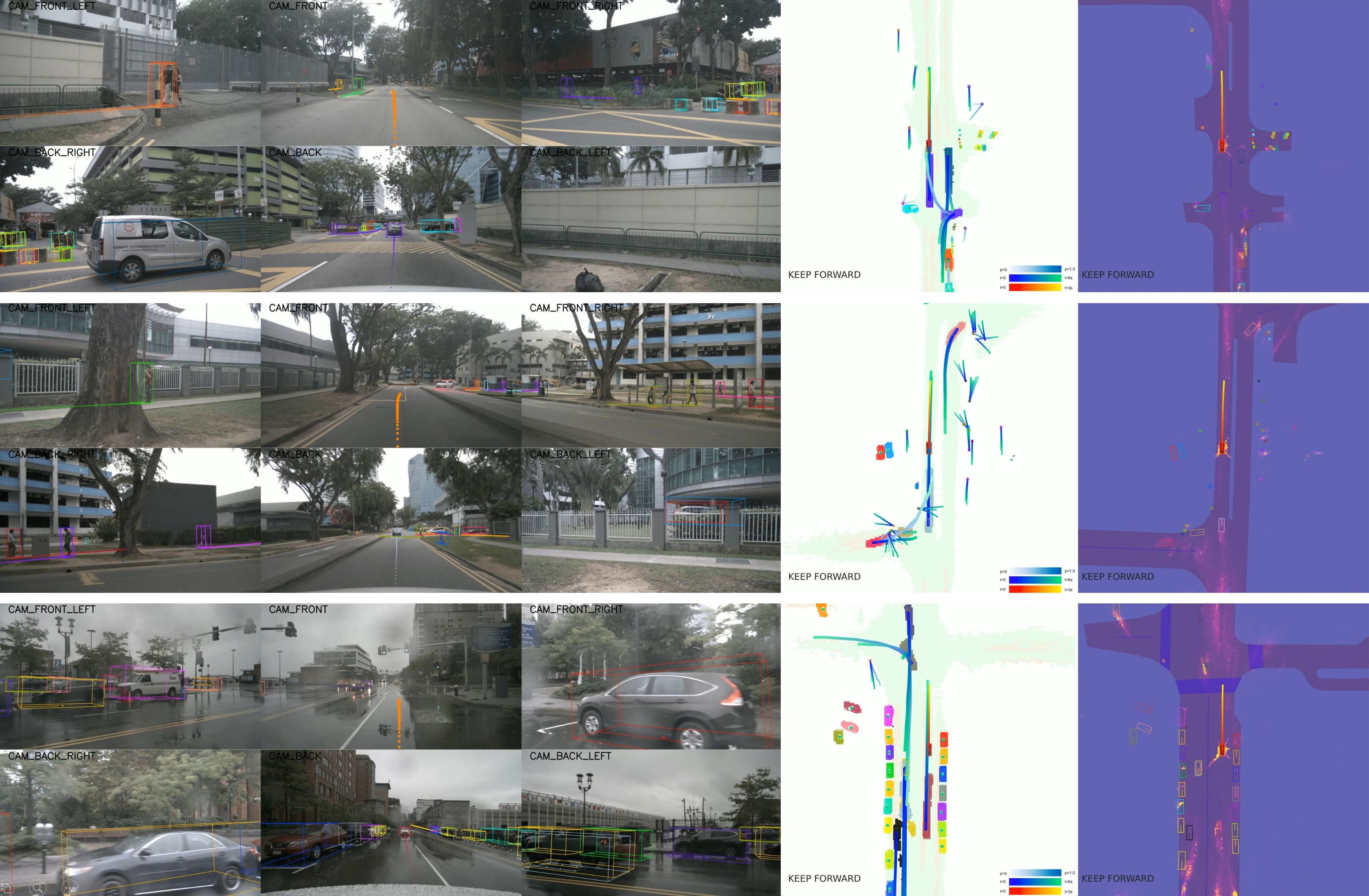}
  \caption{\textbf{Visualization for cruising around the urban areas.} \algname can generate high-quality interpretable perceptual and prediction results, and make a safe maneuver.
  The first three columns show six camera views, and the last two columns are the predicted results and the attention mask from the planning module respectively. Each agent is illustrated with a unique color. Only top-1 and top-3 trajectories from motion forecasting are selected for visualization on images-view and BEV respectively.}
  \label{fig:vis-supp-best}
\end{figure*}

\begin{figure*}[t!]
  \centering
  \includegraphics[width=0.98\textwidth]{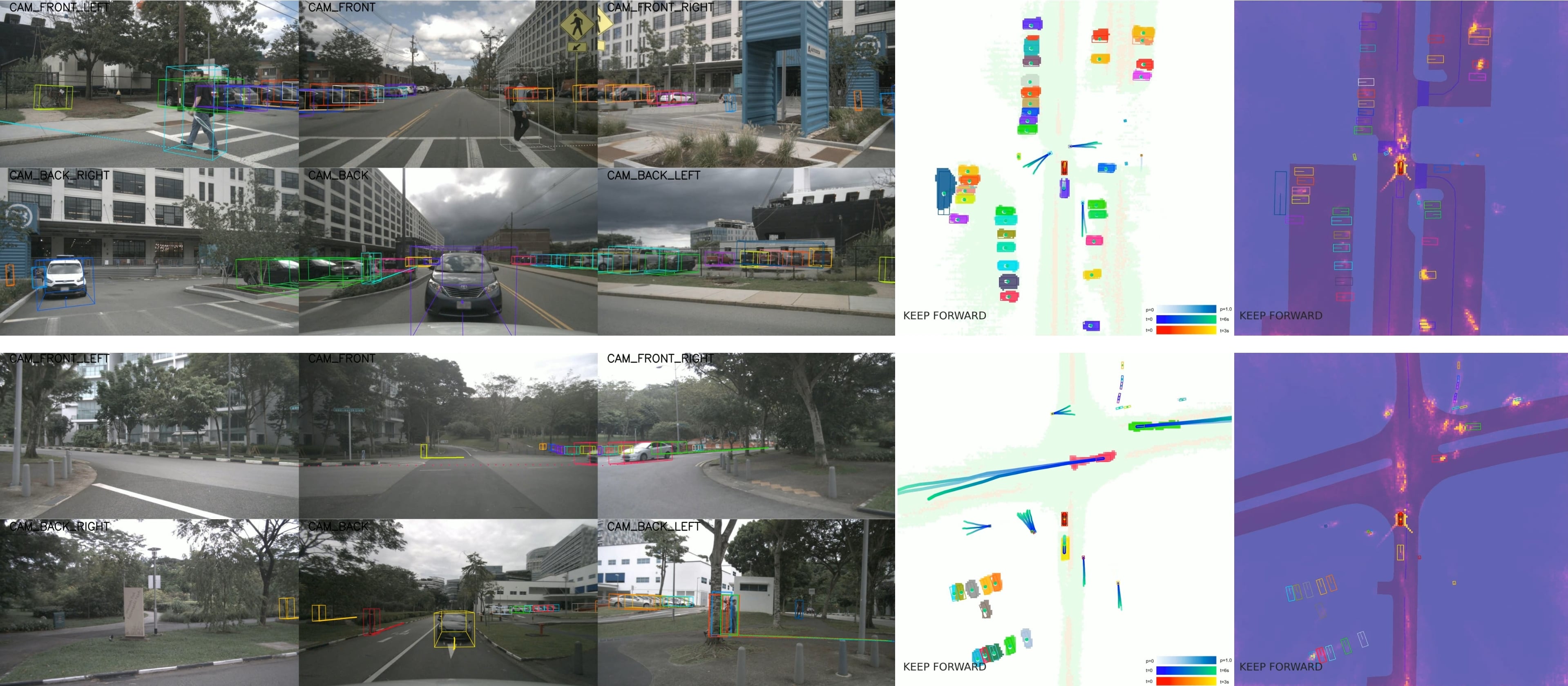}
  \caption{\textbf{Critical case visualization.} Here we demonstrate two critical cases. The first scenario (top) shows that the ego vehicle is yielding to two pedestrians crossing the street, and the second scenario (down) shows that the ego vehicle is yielding to a fast-moving car and waiting to go straight without protection near an intersection. We can observe that much attention is paid to the most critical agents, \ie, pedestrians and fast-moving vehicles, as well as the intended goal location. }
  \label{fig:vis-supp-critical}
\end{figure*}

\begin{figure*}[t!]
  \centering
  \includegraphics[width=\textwidth]{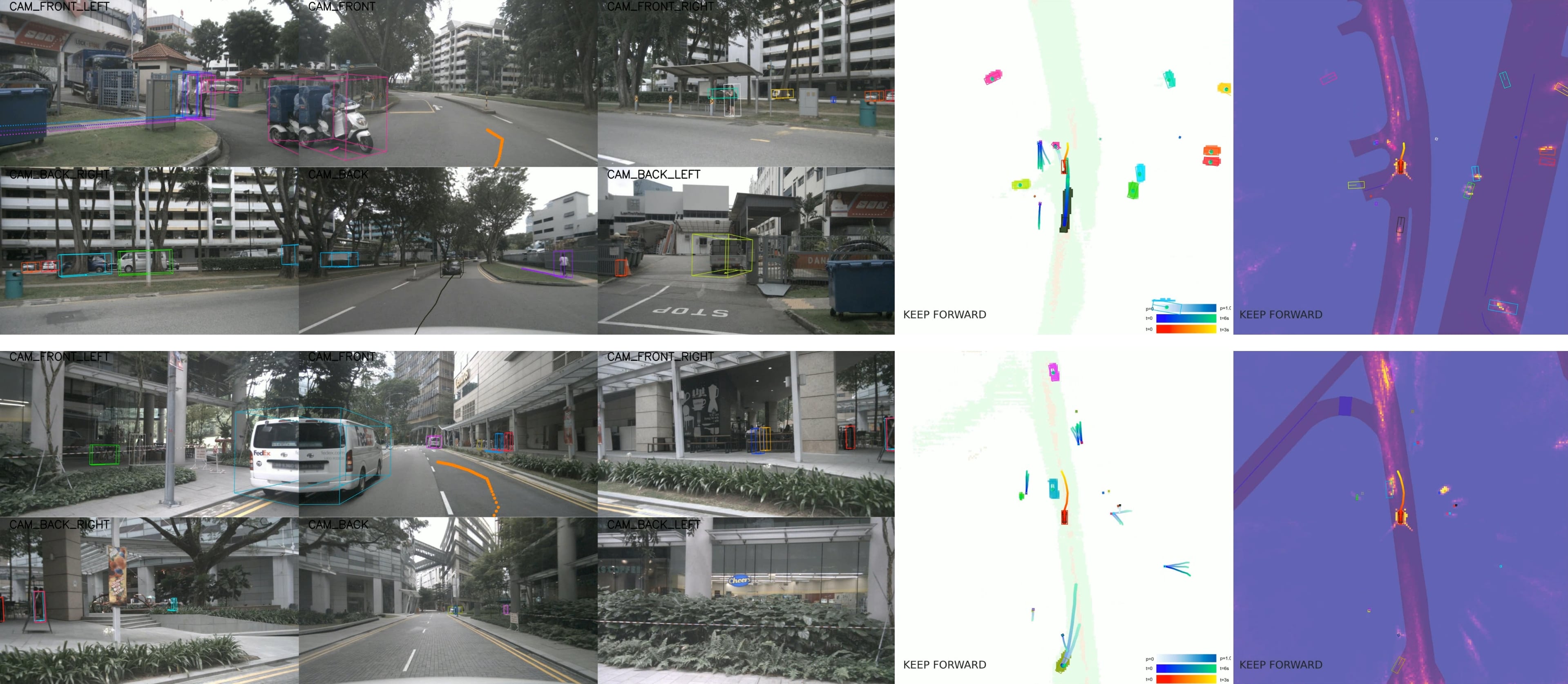}
  \vspace{-12pt}
  \caption{\textbf{Obstacles avoidance visualization.}
  In these two scenarios, the ego vehicle is changing lanes attentively to avoid the obstacle vehicle. From the attention mask, we can observe that our method focuses on the obstacles as well as the road in the front and back.}
  \label{fig:vis-planning-obs-avoid}
\end{figure*}

\begin{figure*}[t!]
  \centering
  \includegraphics[width=\textwidth]{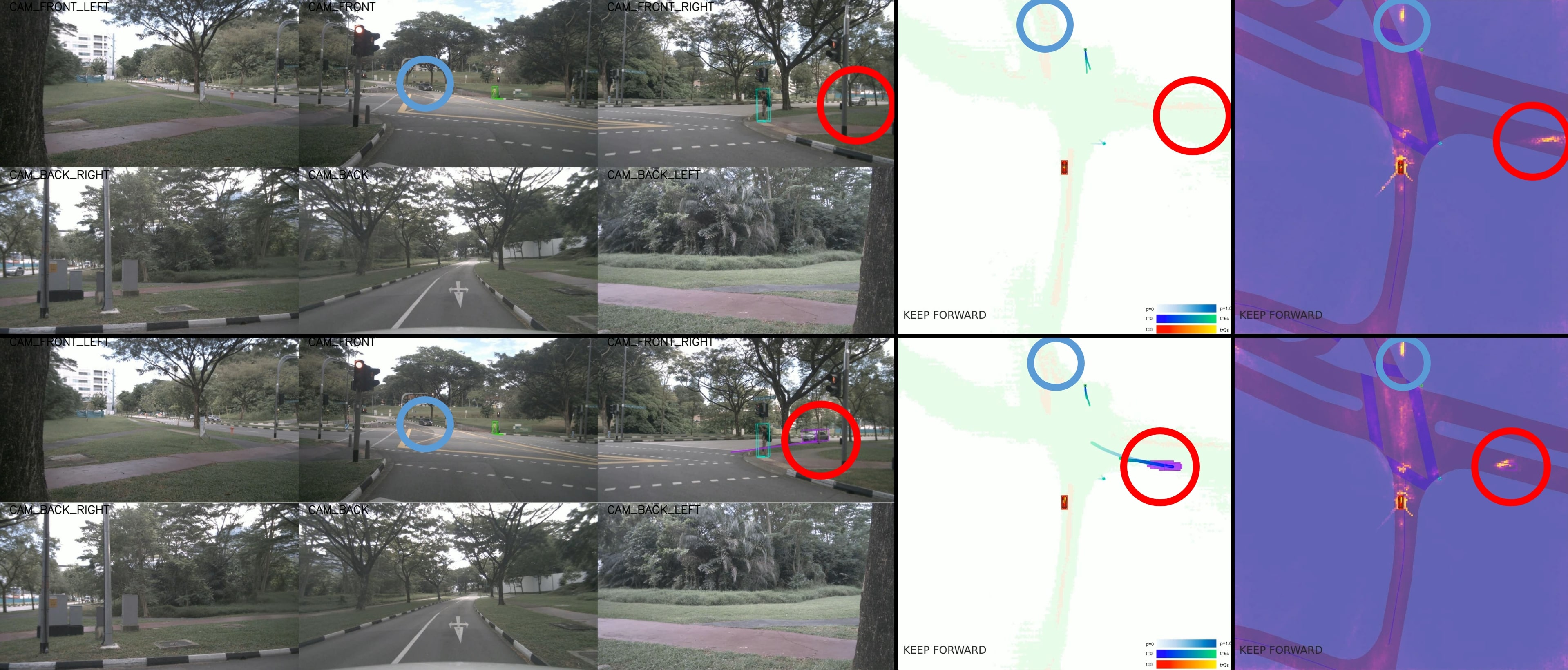}
  \vspace{-12pt}
  \caption{\textbf{Visualization for planning recovering from perception failures.} We show an interesting case where inaccurate results occur in prior modules while the later tasks could still recover. The top row and the down row represent two consecutive frames from the same scenario.
  The vehicle in the red circle is moving from a far distance toward the intersection at a high speed. It is observed that the tracking module misses it at first, then captures it at the latter frame.
  The blue circles show a stationary car yielding to the traffic, and it is missed in both frames. Interestingly, both vehicles show strong reactions to the attention masks of the planner, even though they are missed in the prior modules. It means that our planner still pays attention to those critical though missed agents, which is intractable in previous fragmented and non-unified driving systems, and demonstrates the robustness of \algname.  }
  \label{fig:vis-supp-failure1}
\end{figure*}

\paragraph{Visualization of different scenarios.}
We provide visualizations for more scenarios, including cruising around the urban areas (\cref{fig:vis-supp-best}), critical cases (\cref{fig:vis-supp-critical}), and obstacle avoidance scenarios (\cref{fig:vis-planning-obs-avoid}). 
One promising evidence for our planning-oriented design is shown in \cref{fig:vis-supp-failure1}, where inaccurate results occur in prior modules while the later tasks could still recover.
Similarly, we show results for all tasks in surround-view images, BEV, as well as the attention mask from the planner.
A demo video\footnote{\url{https://opendrivelab.github.io/UniAD/}} is also provided for reference.

\begin{figure*}[t!]
  \centering
  \includegraphics[width=\textwidth]{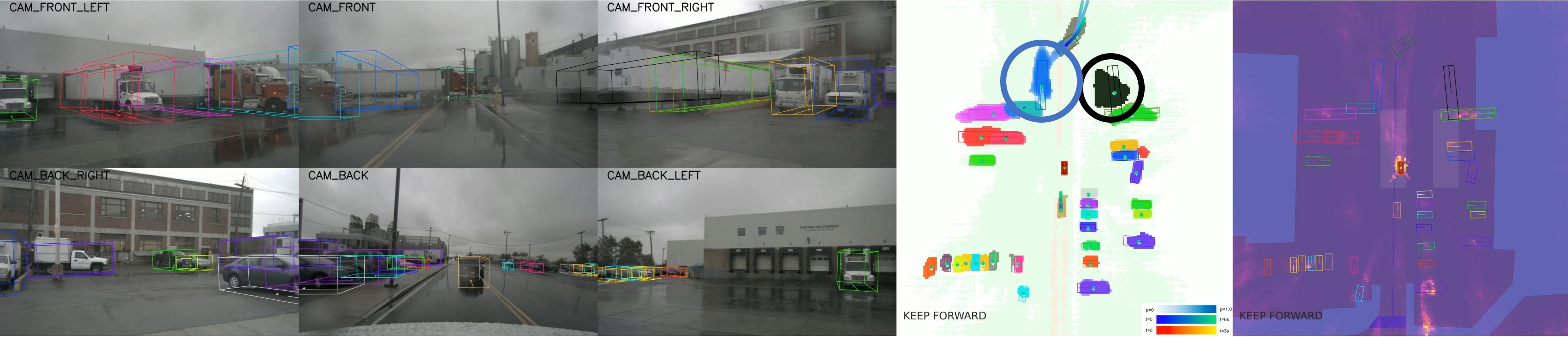}
  \vspace{-12pt}
  \caption{\textbf{Failure cases 1.} Here we present a long-tail scenario, where a large trailer with a white container occupies the entire road. We can observe that our tracking module fails to detect the accurate size of the front trailer and heading angles of vehicles beside the road.
  }
  \label{fig:vis-supp-failure2}
\end{figure*}

\begin{figure*}[t!]
  \centering
  \includegraphics[width=\textwidth]{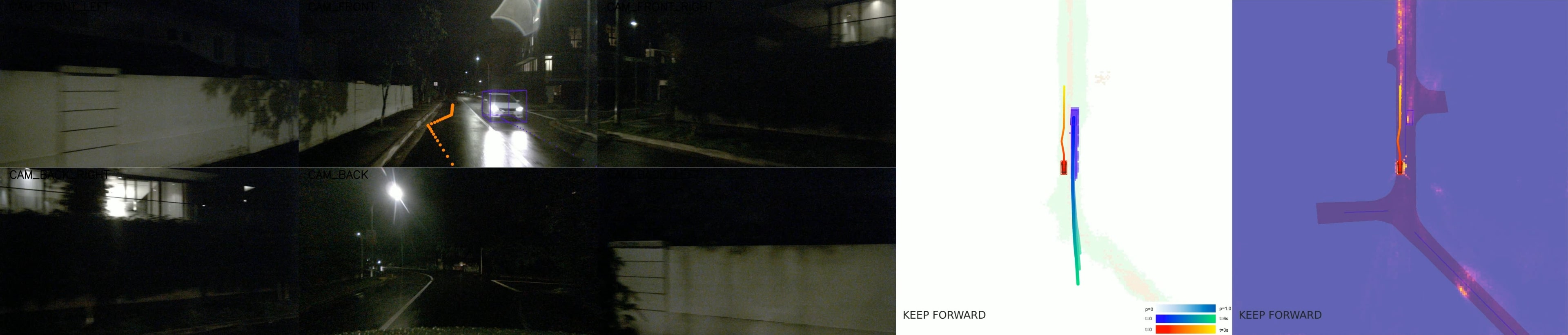}
  \vspace{-12pt}
  \caption{\textbf{Failure cases 2.} In this case, the planner is over-cautious about the incoming vehicle in the narrow street. The dark environment is one critical type of long-tail scenarios in autonomous driving. Applying smaller collision loss weight and more regulation regarding the boundary might mitigate the problem.}
  \label{fig:vis-supp-failure3}
\end{figure*}

\paragraph{Failure cases} are essential for an autonomous driving algorithm to understand its weakness and guide future work, and here we present some failure cases of \algname.
The failure cases of \algname are mainly under some long-tail scenarios where all modules are affected, as depicted in~\cref{fig:vis-supp-failure2} and~\cref{fig:vis-supp-failure3}.

\end{document}